\documentclass[twoside]{article}
\usepackage[utf8]{inputenc}
\usepackage{amsmath}
\usepackage{amsthm}
\usepackage{amsfonts}
\usepackage{amssymb}
\usepackage{float}
\usepackage{graphicx}
\usepackage{subfigure}
\usepackage[left=2cm,right=2cm,top=2cm,bottom=2cm]{geometry}
\usepackage{bm}
\usepackage{color}
\newenvironment{Proof}{\noindent{\sc Proof.}}{\qed}
\newtheorem{theorem}{Theorem}[section]
\newtheorem{lemma}{Lemma}[section]
\newtheorem{cor}{Corollary}[section]
\newtheorem{definition}{Definition}[section]
\newtheorem{prop}{Proposition}[section]
\newtheorem{uda}{Example}[section]
\newtheorem{rem}{Remark}[section]
\renewcommand{\theequation}{\arabic{section}.\arabic{equation}}
\def\bhag#1{\noindent
\setcounter{equation}{0}
\section{#1}
}

\def\HH{{\mathbb H}}

\def\RR{{\mathbb R}}
\def\CC{{\mathbb C}}
\def\ZZ{{\mathbb Z}}

\def\PPI{{{\rm I}\kern-1pt\Pi}}
\def\SS{{\mathbb S}}

\def\TT{\mathbb T}

\def\b #1;{{\bf #1}}
\def\x{{\bf x}}
\def\k{{\bf k}}
\def\y{{\bf y}}

\def\w{{\bf w}}
\def\z{{\bf z}}

\def\O{{\cal O}}

\def\C{{\mathcal C}}
\def\esssup{\mathop{\hbox{{\rm ess sup}}}}

\def\be{\begin{equation}}
\def\ee{\end{equation}}
\def\bea{\begin{eqnarray}}
\def\eea{\end{eqnarray}}
\def\eref#1{(\ref{#1})}
\def\disp{\displaystyle}

\def\donchitre#1#2{\vskip 6.5cm\noindent
\parbox[t]{1in}{\special{eps:#1.eps x=6.5cm y=5.5cm}}
\hbox to 7cm{}\parbox[t]{0.0cm}{\special{eps:#2.eps x=6.5cm y=5.5cm}}}

\def\tn{|\!|\!|}
\def\XX{{\mathbb X}}
\def\BB{{\mathbb B}}
\def\span{\mbox{{\rm span }}}
\def\bs#1{{\boldsymbol{#1}}}

\providecommand{\esssup}{\operatorname*{ess\ sup}}

\title{Kernel based analysis of  massive data}
\author{
H.~N.~Mhaskar\thanks{Institute of Mathematical Sciences, Claremont Graduate University, Claremont, CA 91711, U.S.A. 
\textsf{email:} hrushikesh.mhaskar@cgu.edu }}
\date{}
\begin{document}
\maketitle
\begin{abstract}
Dealing with massive data is a challenging task for machine learning.
An important aspect of machine learning is function approximation.
In the context of  massive data, some of the commonly used tools for this purpose are sparsity, divide-and-conquer, and distributed learning. 
In this paper, we develop a very general theory of approximation by networks, which we have called eignets, to achieve local, stratified approximation.
The very massive nature of the data allows us to use these eignets to solve inverse problems such as finding a good approximation to the probability law that governs the data, and finding  the local smoothness of the target function near different points in the domain. 
In fact, we develop a wavelet-like representation using our eignets.
Our theory is applicable to approximation on a general locally compact metric measure space.
Special examples include approximation by  periodic basis functions on the torus, zonal function networks on a Euclidean sphere (including smooth ReLU networks),  Gaussian networks, and approximation on manifolds.
We construct pre-fabricated networks so that no data-based training is required for the approximation.
\end{abstract}

%Input $f$\\
%
%$\{\hat{f}(k)\}$\\
%
%$x$\\
%
%$T$\\
%
%$\{\hat{u}(k)(t)\}$\\
%
%$\{\hat{u}(k)(T)\}$\\
%
%$u(x,T)$\\
%
%Preprocessing\\
%
%Deep network\\
%
%Shallow network

%$t<T$

\bhag{Introduction}\label{bhag:intro}

Rapid advances in technology have led to the availability and need to analyze a massive data. 
The problem arises in almost every area of life from medical science to homeland security to finance.
An immediate problem in dealing with a massive data set is that it is not possible to store it in a computer memory, so that one has to deal with the data piecemeal to keep access to an external memory to a minimum.
The other challenge is to devise efficient numerical algorithms to overcome difficulties, for example, in  using the customary optimization problems in machine learning.
On the other hand, the very availability of a massive data set should lead also to opportunities to solve some problems here-to-fore considered unmanageable. 
For example, deep learning often requires a large amount of training data, which in turn, helps to figure out the granularity in the data.
Apart from deep learning, distributed learning is also a popular way of dealing with big data.
A good survey with the taxonomy for dealing with massive data is given recently in \cite{zhou2017machine}.

As pointed out in \cite{girosi1990networks, CucSma02, zhoubk_learning}, the main task in machine learning can be viewed as one of approximation of functions based on noisy values of the target function, sampled at points which are themselves sampled from an unknown distribution.
Therefore, it is natural to seek approximation theory techniques to solve the problem.
However, most of the classical approximation theory results are either not constructive, or else study function approximation only on known domains.
In this century, there is a new paradigm to consider function approximation on data-defined manifolds; a good introduction to the subject is in the special issue \cite{achaspissue} of Applied and Computational Harmonic Analysis, edited by Chui and Donoho.
In this theory, one assumes the \emph{manifold hypothesis}, i.e.,  that the data is sampled from a probability distribution $\mu^*$ supported on a smooth, compact,  connected, Riemannian  manifold; for simplicity, even that $\mu^*$ is the Riemannian volume measure for the manifold, normalized to be a probability measure.
Following, e.g., \cite{belkin2003laplacian, belkinfound, niyogi2, lafon, singer}, one constructs first a ``graph Laplacian'' from the data, and finds its eigen-decomposition. 
It is proved in the above mentioned papers that as the size of the data tends to infinity, the graph Laplacian converges to the Laplace-Beltrami operator on the manifold and the eigen-values (respectively, eigen-vectors) converge to the corresponding quantities on the manifold.
A great deal of work is devoted to studying the geometry of this unknown manifold (e.g., \cite{jones2010universal, liao2016adaptive}), based on the so called heat kernel.
The theory of function approximation on such manifolds is also well developed (e.g., \cite{mauropap, eignet, heatkernframe, compbio, modlpmz}).

A bottleneck in this theory is the computation of the eigen-decomposition of a matrix, which is necessarily huge in the case of big data.
Kernel based methods have been used also in connection with approximation on manifolds (e.g. \cite{rosasco2010learning, rudi2017falkon, sergeibk, data_based_construction2019,sergei_multiple_kernels2017}).
The kernels used in this method are constructed typically as a radial basis function (RBF) in the ambient space, and the methods are traditional machine learning methods involving optimization.
As mentioned earlier, massive data poses a big challenge for the solution of these optimization problems.
The theoretical results in this connection assume a Mercer's expansion in terms of the Laplacian eigenfunctions for the kernel, satisfying certain conditions. 
In this paper, we develop a general theory including several RBF kernels in use in different contexts (examples are discussed in Section~\ref{bhag:motive}). 
Rather than using optimization based techniques, we will provide a direct construction of the approximation, based on what we have called eignets. 
An eignet is defined directly using the eigen-decomposition on the manifold.
Thus, we focus directly on the properties of Mercer expansion in an abstract and unified manner that enables us to construct local approximations suitable for working with massive data without using optimization.

It is also possible that the manifold hypothesis does not hold, and there is a recent work \cite{fefferman2016testing} by Fefferman, Mitter, and Narayanan proposing an algorithm to test this hypothesis.
On the other hand, our theory for function approximation does not necessarily use the full strength of Riemannian geometry.
In this paper, we have therefore decided to work with a general locally compact metric measure space, isolating those properties which are needed for our analysis, and substituting some which are not applicable in the current setting.

Our motivation comes from some recent works on distributed learning by Zhou and his collaborators \cite{chui2019realization, guo2017learning, lin2019distributed} as well as our own work on deep learning \cite{dingxuanpap, mhaskar2019function}. 
For example, in \cite{lin2019distributed}, the approximation is done on the Euclidean sphere using a localized kernel introduced in \cite{locsmooth}, where the massive data is divided into smaller parts, each dense on the sphere, and the resulting polynomial approximations are added to get the final result.
In \cite{chui2019realization}, the approximation takes place on a cube, and exploits any known sparsity in the representation of the target function in terms of spline functions.
In \cite{dingxuanpap, mhaskar2019function}, we have argued that from a function approximation point of view, the observed superiority of deep networks over shallow ones results from the ability of deep networks to exploit any compositional structure in the target function. 
For example, in image analysis, one may divide the image into smaller patches, which are then combined in a hierarchical manner, resulting in a tree structure \cite{smale2010mathematics}. 
By putting a shallow network at each node to learn those aspects of the target function which depend upon the pixels seen up to that level, one can avoid the curse of dimensionality.
In some sense, this is a divide-and-conquer strategy, not so much on the data set itself, but on the dimension of the input space.

The highlights of this paper are the following.
\begin{itemize}
\item In order to avoid an explicit, data dependent eigen-decomposition,  we introduce the notion of an eignet, that generalizes several radial basis function and zonal function networks. 
We construct pre-fabricated eignets, whose linear combinations can be constructed just by using the noisy values of the target function as the coefficients, to yield the desired approximation.
\item
Our theory generalizes the results in a number of examples used commonly in machine learning, some of which we will describe in Section~\ref{bhag:motive}.
\item 
The use of optimization methods such as empirical risk minimization has an intrinsic difficulty, namely, the minimizer of this risk may have no connection with the approximation error. 
Also, there are other problems such as local minima, saddle points, speed of convergence, etc. that need to be taken into account, and the massive nature of the data makes this into an even more challenging task.
 Our results do not depend upon any kind of  optimization in order to determine the necessary approximation.
\item We develop a theory for local approximation using eignets, so that only a relatively small amount of data is used in order to approximate the target function in any ball of the space, the data sub-sampled using a distribution supported on a neighborhood of that ball. 
The accuracy of approximation adjusts itself automatically depending upon the local smoothness of the target function on the ball.
\item In usual machine learning algorithms, it is customary to assume a prior on the target function, called smoothness class in approximation theory parlance. 
Our theory demonstrates clearly how a massive data can actually help to solve the inverse problem to determine the local smoothness of the target function using a wavelet-like representation, based solely on the data.
\item Our results allow one to solve the inverse problem of estimating the probability density from which the data is chosen.
In contrast to the statistical approaches that we are aware of, there is no limitation on how accurate the approximation can be asymptotically in terms of the number of samples; the accuracy is determined entirely by the smoothness of the density function.
\item All our estimates are given in terms of probability of the error being small, rather than the expected value of some loss function being small.
\end{itemize}

Necessarily, the paper is very abstract, theoretical, and technical. 
In Section~\ref{bhag:motive}, we present a number of examples which are generalized by our set-up.
The abstract set-up, together with the necessary definitions and assumptions are discussed in Section~\ref{bhag:setup}.
The main results are stated in Section~\ref{bhag:mainresults} and proved in Section~\ref{bhag:pfsect}. 
The proofs require a great deal of preparation, which is presented in Sections~\ref{bhag:ddrprep}, \ref{bhag:locapprox}, and \ref{bhag:quadrature}. 
The results in these sections are not all new. 
Many of them are new only in some nuance. 
For example, we have proved in Section~\ref{bhag:quadrature} the quadrature formulas required in the construction of our pre-fabricated networks in a probabilistic setting, and substituting an estimate on the gradients by certain Lipschitz condition, which makes sense without the differentiability structure on the manifold as we had done in our previous works.
Our Theorem~\ref{theo:mztheo} generalizes most of our previous results in this direction, except for \cite[Theorem~2.3]{bitrep}.
We have strived to give as many proofs as possible, partly for the sake of completion and partly because the results were not stated earlier in exactly the same form as needed here.
In Appendix~\ref{bhag:gaussupbd}, we give a short proof of the fact that the Gaussian upper bound for the heat kernel holds for arbitrary smooth, compact, connected manifolds.
We could not find a reference for this fact.
In Appendix~\ref{bhag:concentration}, we state the main probability theory estimates that are used ubiquitously in the paper.

\bhag{Motivating examples}\label{bhag:motive}
In this paper, we aim to develop a unifying theory applicable to a variety of kernels and domains.
In this section, we describe some examples which have motivated the abstract theory to be presented in the rest of the paper.
In the following examples, $q\ge 1$ is a fixed integer.
\begin{uda}\label{uda:torus}
{\rm
Let $\TT^q=\RR^q/(2\pi\ZZ^q)$ be the $q$-dimensional torus. 
The distance between points $\x=(x_1,\cdots,x_q)$ and $\y=(y_1,\cdots, y_q)$ is defined by $\max_{1\le k\le q}|(x_k-y_k)\mbox{ mod } 2\pi|$. 
The trigonometric monomial system $\{\exp(i\k\cdot\circ) : \k\in\ZZ^q\}$ is orthonormal with respect to the Lebesgue measure normalized to be a probability measure on $\TT^q$. 
We recall that the periodization of a function $f :\RR^q\to \RR$ is defined formally by $f^\circ(\x)=\sum_{\k\in\ZZ^q}f(\x+2\k \pi)$.
When $f$ is integrable then the Fourier transform of $f$ at $\k\in\ZZ^q$ is the same as the $\k$-th Fourier coefficient of $f^\circ$. This Fourier coefficient will be denoted by $\widehat{f^\circ}(\k)=\hat{f}(\k)$.
 A periodic basis function network has the form $\x\mapsto\sum_{k=1}^n a_kG(\x-\x_k)$, where $G$ is a periodic function, called the activation function. 
The examples of the activation functions in which we are interested in this paper include:
\begin{enumerate}
\item  Periodization of the Gaussian.
$$
G(\x) = \sum_{\k \in \ZZ^q} \exp(-|\x-2\pi\k|_2^2 /2)~, \hat{G} (\k) = (2 \pi)^{q/2} \exp(-|\k|_2^2 / 2).
$$
\item  Periodization of the Hardy multiquadric.\footnote{A
Hardy multiquadric is a function of the form $\x\to
(\alpha^2+|\x|_2^2)^{-1}$, $\x\in\RR^q$. It is one of the oft--used
function in theory and applications of radial basis function networks.
For a survey, see the paper \cite{hardy1990theory} of Hardy.}
$$
G(\x)= \sum_{\k\in\ZZ^q} (\alpha^2+|\x-2\pi\k|_2^2)^{-1}~, \quad \hat{G}(\k) = \frac{\pi^{(q+1)/2}}{\Gamma \left( \frac{q+1}{2} \right) \alpha} \exp(-\alpha |\k|_2), \qquad \alpha>0.\qquad\qquad \qed
$$
\end{enumerate}
}
\end{uda}

\begin{uda}\label{uda:cube}
{\rm
If $\x=(x_1,\cdots,x_q)\in [-1,1]^q$, there exists a unique $\bs\theta=(\theta_1,\cdots,\theta_q)\in [0,\pi]^q$ such that $\x=\cos(\bs\theta)$. 
Therefore, $[-1,1]^q$ can be thought of as a quotient space of $\TT^q$ where all points of the form $\bs\varepsilon\odot\bs\theta=\{(\varepsilon_1\theta_1,\cdots,\varepsilon_q\theta_q)\}$, $\bs\varepsilon=(\varepsilon_1,\cdots,\varepsilon_q)\in \{-1,1\}^q$, are identified.
Any function on $[-1,1]^q$ can then by lifted to $\TT^q$, and this lifting preserves all the smoothness properties of the function.
Our set-up below includes $[-1,1]^q$, where the distance and the measure are defined via the mapping to the torus, and suitably weighted Jacobi polynomials are considered to be the orthonormalized family of functions.
In particular, if $G$ is a periodic activation function, $\x=\cos(\bs\theta)$, $\y=\cos(\bs\phi)$, then the function $G^\square(\x,\y)=\sum_{\bs\varepsilon\in \{-1,1\}^q}G(\bs\varepsilon\odot(\bs\theta-\bs\phi))$ is an activation function on $[-1,1]^q$ with an expansion $\sum_{\k\in\ZZ_+^q}b_\k T_\k(\x) T_\k(\y)$, where $T_\k$'s are tensor product, orthonormalized, Chebyshev polynomials.
Furthermore, $b_\k$'s have the same asymptotic behavior as $\hat{G}(\k)$'s.
\qed}
\end{uda}

\begin{uda}\label{uda:sphere}
{\rm
Let $\SS^q=\{\x\in\RR^{q+1}: |\x|_2=1\}$ be the unit sphere in $\RR^{q+1}$. 
The dimension of $\SS^q$ as a manifold is $q$.
We assume the geodesic distance $\rho$ on $\SS^q$, and the volume measure $\mu^*$ normalized to be a probability measure.
We refer the reader to \cite{mullerbk} for details, describing here only the essentials to get a ``what-it-is-all-about'' introduction.
The set of (equivalence classes) of restrictions to polynomials in $q+1$ variables with total degree $<n$ to $\SS^q$ are called spherical polynomials of degree $<n$. 
The set of restrictions of homogeneous harmonic polynomials of degree $\ell$ to $\SS^q$ is denoted by $\HH_\ell$, with dimension $d_\ell$. 
There is an orthonormal basis $\{Y_{\ell,k}\}_{k=1}^{d_\ell}$ for each $\HH_\ell$, that satisfies an addition formula
$$
 \sum_{k=1}^{d_\ell} Y_{\ell,k}(\b x;){Y_{\ell,k}(\b y;)} = \omega_{q-1}^{-1}p_\ell(1)p_\ell(\x\cdot\y),
$$
where $\omega_{q-1}$ is the volume of $\SS^{q-1}$, and $p_\ell$ is the degree $\ell$ ultraspherical polynomial so that the family $\{p_\ell\}$ is orthonormalized with respect to the measure $(1-x^2)^{(q-2)/2}$ on $(-1,1)$.
A zonal function  on the sphere has the form $\x\mapsto G(\x\cdot\y)$, where the activation function $G :[-1,1]\to\RR$ has a formal expansion of the form
$$
G(t)=\omega_{q-1}^{-1}\sum_{\ell=0}^\infty \hat{G}(\ell)p_\ell(1)p_\ell(t).
$$
In particular, formally, $G(\x\cdot\y)= \sum_{\ell=0}^\infty \hat{G}(\ell) \sum_{k=1}^{d_\ell}Y_{\ell,k}(\b x;){Y_{\ell,k}(\b y;)}$. 
The examples of the activation functions in which we are interested in this paper include:
\begin{enumerate}
\item 
$$
G_r(x) := (1-2rx+r^2)^{-(q-1)/2}, \qquad x\in [-1,1], \ 0<r<1.
$$
It is shown in \cite[Lemma 18]{mullerbk} that
$$
\widehat{G_r}(\ell)= \frac{(q-1)\omega_q}{2\ell+q-1}r^\ell, \qquad \ell=1,2,\cdots.
$$
\item $$
G_r^E(x) := \exp(rx), \qquad x\in [-1,1], \ r>0.
$$
It is shown in \cite[Lemma~5.1]{mnw2} that
$$
\hat{G_r^E}(\ell) = \frac{\omega_q r^\ell}{2^\ell\,\Gamma(\ell+\frac{q+1}2)}
\biggl(1+\O(1/\ell)\biggr).
$$
\item The smooth ReLU function $G(t)= \log(1+e^t)=t_++\O(e^{-|t|})$.
The function $G$ has an analytic extension to the strip $\RR+(-\pi,\pi)i$ of the complex plane. 
So, Bernstein approximation theorem \cite[Theorem~5.4.2]{timanbk} can be used to show that
$$
\limsup_{\ell\to\infty}|\hat{G}(\ell)|^{1/\ell}=1/\pi. \qquad\qquad \qed
$$
\end{enumerate}
}
\end{uda}

\begin{uda}\label{uda:genmanifold}
{\rm 
Let $\XX$ be a smooth, compact, connected Riemannian manifold (without boundary), $\rho$ be the geodesic distance on $\XX$, $\mu^*$ be the Riemannian volume measure normalized to be a probability measure, $\{\lambda_k\}$ be the sequence of eigenvalues of the (negative) Laplace-Beltrami operator on $\XX$, and $\phi_k$ be the eigenfunction corresponding to the eigenvalue $\lambda_k$; in particular, $\phi_0\equiv 1$.
This example, of course, includes Examples~\ref{uda:torus}, \ref{uda:cube}, and \ref{uda:sphere}.
An eignet in this context has the form $x\mapsto \sum_{k=1}^n a_kG(x,x_k)$, where the activation function $G$ has a formal expansion of the form $G(x,y)=\sum_k b(\lambda_k)\phi_k(x)\phi_k(y)$.
One interesting example is the heat kernel:
$
\disp\sum_{k=0}^\infty \exp(-\lambda_k^2t)\phi_k(x)\phi_k(y)
$.
\qed}
\end{uda}

\begin{uda}\label{uda:genhermite}
{\rm 
Let $\XX=\RR^q$, $\rho$ be the $\ell^\infty$ norm on $\XX$, $\mu^*$ be the Lebesgue measure. 
For any multi-integer $\k\in\ZZ^q_+$, the (multivariate) Hermite function $\phi_\k$ is defined via the generating function 
\be\label{hermitegenfunc}
\sum_{\k\in\ZZ^q_+}\frac{\phi_\k(\x)}{\sqrt{2^{|\k|_1} \k!}}\w^\k=\pi^{-1/4}\exp\left(-\frac{1}{2}|\x-\w|_2^2+|\w|_2^2/4\right), \qquad \w\in \CC^q.
\ee
The system $\{\phi_\k\}$ is orthonormal with respect to $\mu^*$, and satisfies
$$
\Delta\phi_\k(\x)-|\x|_2^2\phi_\k(\x)=-(2|\k|_1+1)\phi_k(\x),\qquad \x\in\RR^q,
$$
where $\Delta$ is the Laplacian operator.
As a consequence of the so called Mehler identity, one obtains (\cite{chuigaussian}) that
\be\label{gaussianexpansion}
\exp\left(-|\x-\frac{\sqrt{3}}{2}\y|_2^2\right)\exp(-|\y|_2^2/4)=\left(\frac{3}{2\pi}\right)^{-q/2}\sum_{\k\in\ZZ_+^d}\phi_\k(\x)\phi_\k(\y)3^{-|\k|_1/2}.
\ee
A Gaussian network is a network of the form $\x\mapsto\sum_{k=1}^n a_k\left(-|\x-\z_k|_2^2\right)$, where it is convenient to think of $\disp\z_k=\frac{\sqrt{3}}{2}\y_k$. 
\qed}
\end{uda}

\bhag{The set-up and definitions}\label{bhag:setup}

\subsection{Data spaces}\label{bhag:dataspace}
Let $\XX$ be a connected, locally compact, metric space with metric $\rho$. For $r>0$, $x\in\XX$, we denote
$$
\mathbb{B}(x,r)=\{y\in\XX : \rho(x,y)\le r\}, \ \Delta(x,r)=\mathsf{closure}(\XX\setminus \mathbb{B}(x,r)).
$$
If $K\subseteq \XX$ and $x\in\XX$, we write as usual $\rho(K,x)=\inf_{y\in K}\rho(y,x)$. 
It is convenient to denote the set\\
 $\{x\in\XX; \rho(K,x)\le r\}$ by $\mathbb{B}(K,r)$.
The diameter of $K$ is defined by $\mathsf{diam}(K)=\sup_{x,y\in K}\rho(x,y)$.

For a Borel measure $\nu$ on $\XX$ (signed or positive), we denote by $|\nu|$ its total variation measure, defined for Borel subsets $K\subset \XX$ by
$$
|\nu|(K)=\sup_{\mathcal{U}}\sum_{U\in\mathcal{U}}|\nu(U)|,
$$
where the supremum is over all countable measurable partitions $\mathcal{U}$ of $K$.
In the sequel, the term measure will mean a signed or positive, complete, sigma-finite Borel measure.
Terms such as measurable will mean Borel measurable.
If $f :\XX\to\RR$ is  measurable,  $K\subset \XX$ is measurable, and $\nu$ is a measure, we define\footnote{$|\nu|\!\!-\!\!\esssup_{x\in K}|f(x)|=\inf\left\{t : |\nu|\left(\{x\in K : |f(x)|>t\}\right)=0\right\}$}
$$
\|f\|_{p,\nu,K}=\begin{cases}
\disp \left\{\int_K |f(x)|^pd|\nu|(x)\right\}^{1/p}, & \mbox{ if $1\le p<\infty$},\\[2ex]
\disp |\nu|\!\!-\!\!\esssup_{x\in K}|f(x)|, & \mbox{ if $p=\infty$}.
\end{cases}
$$
The symbol $L^p(\nu,K)$ denotes the set of all measurable functions $f$ for which $\|f\|_{p,\nu,K}<\infty$, with the usual convention that two functions are considered equal if they are equal $|\nu|$-almost everywhere on $K$. 
The set $C_0(K)$ denotes the set of all uniformly continuous  functions on $K$ vanishing at $\infty$.
In the case when $K=\XX$, we will omit the mention of $K$, unless it is necesssary to mention it to avoid confusion.

We fix a non-decreasing sequence $\{\lambda_k\}_{k=0}^\infty$, with $\lambda_0=0$ and $\lambda_k\uparrow\infty$ as $k\to\infty$. 
We also fix  a positive, sigma-finite, Borel measure $\mu^*$ on $\XX$, and a system of orthonormal  functions $\{\phi_k\}_{k=0}^\infty\subset L^1(\mu^*,\XX)\cap C_0(\XX)$, such that $\phi_0(x)>0$ for all $x\in \XX$. 
We define
\be\label{diffpolyclass}
\Pi_n=\span\{\phi_k : \lambda_k <n\}, \qquad n>0.
\ee
It is convenient to write $\Pi_n =\{0\}$ if $n\le 0$ and $\Pi_\infty =\bigcup_{n>0}\Pi_n$.
It will be assumed in the sequel that $\Pi_\infty$ is dense in $C_0$ (and hence, in every $L^p$, $1\le p <\infty$). 
We will often refer to the elements of $\Pi_\infty$ as \textbf{diffusion polynomials} in keeping with \cite{mauropap}.

\begin{definition}\label{def:fastdecreasing}
We will say that a sequence $\{a_n\}$ (or a function $F : [0,\infty)\to\RR$) is \textbf{fast decreasing} if $\disp\lim_{n\to\infty} n^Sa_n =0$ (respectively, $\disp\lim_{x\to\infty}x^Sf(x)=0$) for every $S>0$. 
A sequence $\{a_n\}$ has \textbf{polynomial growth} if there exist $c_1, c_2>0$ such that $|a_n|\le c_1n^{c_2}$ for all $n\ge 1$, and similarly for functions.
\end{definition}

\begin{definition}\label{ddrdef}
The space $\XX$ (more precisely, the tuple $\Xi=(\XX,\rho,\mu^*, \{\lambda_k\}_{k=0}^\infty, \{\phi_k\}_{k=0}^\infty)$) is called a \textbf{data space} if each of the following conditions is satisfied.
\begin{enumerate}
\item For each $x\in\XX$, $r>0$, $\mathbb{B}(x,r)$ is compact.
\item (\textbf{Ball measure condition}) There exist $q\ge 1$ and $\kappa>0$ with the following property: For each $x\in\XX$, $r>0$,
\be\label{ballmeasurecond}
\mu^*(\mathbb{B}(x,r))=\mu^*\left(\{y\in\XX: \rho(x,y)<r\}\right)\le \kappa r^q.
\ee
(In particular, $\mu^*\left(\{y\in\XX: \rho(x,y)=r\}\right)=0$.)
\item (\textbf{Gaussian upper bound}) There exist $\kappa_1, \kappa_2>0$ such that for all $x, y\in\XX$, $0<t\le 1$,
\be\label{gaussianbd}
\left|\sum_{k=0}^\infty \exp(-\lambda_k^2t)\phi_k(x)\phi_k(y)\right| \le \kappa_1t^{-q/2}\exp\left(-\kappa_2\frac{\rho(x,y)^2}{t}\right).
\ee
\item (\textbf{Essential compactness}) For every $n\ge 1$ there exists a compact set $\mathbb{K}_n\subset\XX$ such that the function $n\mapsto \mathsf{diam}(\mathbb{K}_n)$ has polynomial growth, while the functions
$$n\mapsto \sup_{x\in\XX\setminus\mathbb{K}_n}\sum_{\lambda_k<n}\phi_k(x)^2
$$ and
$$
n\mapsto \int_{\XX\setminus\mathbb{K}_n}\left(\sum_{\lambda_k<n}\phi_k(x)^2\right)^{1/2}d\mu^*(x)
$$
are both fast decreasing.
(Necessarily, $n\mapsto\mu^*(\mathbb{K}_n)$ has polynomial growth as well.)
\end{enumerate}
\end{definition}

\begin{rem}\label{rem:kn_nested}
{\rm
We assume without loss of generality that $\mathbb{K}_n \subseteq \mathbb{K}_m$ for all $n<m$ and that $\mu^*(\mathbb{K}_1)>0$.
\qed}
\end{rem}
\begin{rem}\label{rem:defclarification}
{\rm
It is clear that if $\XX$ is compact, then the first condition as well as the essential compactness condition are automatically satisfied. We may take $\mathbb{K}_n =\XX$ for all $n$.
In this case, we will assume tacitly that $\mu^*$ is a probability measure, and $\phi_0\equiv 1$.
\qed}
\end{rem}

\begin{uda}\label{uda:manifold}
{\rm (\textbf{Manifold case})
This example points out that our notion of data space generalizes the set-ups in Examples~\ref{uda:torus}, \ref{uda:cube}, \ref{uda:sphere}, and \ref{uda:genmanifold}. Let $\XX$ be a smooth, compact, connected Riemannian manifold (without boundary), $\rho$ be the geodesic distance on $\XX$, $\mu^*$ be the Riemannian volume measure normalized to be a probability measure, $\{\lambda_k\}$ be the sequence of eigenvalues of the (negative) Laplace-Beltrami operator on $\XX$, and $\phi_k$ be the eigenfunction corresponding to the eigenvalue $\lambda_k$; in particular, $\phi_0\equiv 1$. If the condition \eref{ballmeasurecond} is satisfied, then $(\XX,\rho,\mu^*, 
\{\lambda_k\}_{k=0}^\infty, \{\phi_k\}_{k=0}^\infty)$ is a data space. 
Of course, the assumption of essential compactness is satisfied trivially. (See Appendix~\ref{bhag:gaussupbd} for the Gaussian upper bound.)
\qed}
\end{uda}

\begin{uda}\label{uda:hermite}
{\rm (\textbf{Hermite case})
We illustrate how Example~\ref{uda:genhermite} is included in our definition of a data space.
Accordingly, we assume the set-up as in that example.
For $a>0$, let $\phi_{\k,a}(x)=a^{-q/2}\phi_\k(a\x)$. 
With $\lambda_\k=\sqrt{|\k|_1}$, the system $\Xi_a=(\RR^q, \rho, \mu^*, \{\lambda_\k\}, \{\phi_{\k,a}\})$ is a data space. When $a=1$, we will omit its mention from the notation in this context.
The first two conditions are obvious.
The Gaussian upper bound follows by the multivariate Mehler identity \cite[Equation~(4.27)]{hermite_recovery}.
The assumption of essential compactness is satisfied with $\mathbb{K}_n=\mathbb{B}(\bs 0, cn)$ for a suitable constant $c$ (cf. \cite[Chapter~6]{mhasbk}.) 
\qed}
\end{uda}

In the rest of this paper, we assume $\XX$ to be a data space. 
Different theorems will require some additional assumptions, two of which we now enumerate. 
Not every theorem will need all of these; we will state explicitly which theorem uses which assumptions, apart from $\XX$ being a data space.

The first of these deals with the product of two diffusion polynomials.
We do not know of any situation where it is not satisfied, but are not able to prove it in general.

\begin{definition}\label{def:prodassumption}
(\textbf{Product assumption})
There exists $A^*\ge 1$ and a family $\{R_{j,k,n}\in\Pi_{A^*n}\}$ such that for every $S>0$,
\be\label{weakprodass}
\lim_{n\to \infty}n^S\left(\max_{\lambda_k,\lambda_j <n,\ p=1,\infty}\|\phi_k\phi_j-R_{j,k,n}\phi_0\|_p\right) =0.
\ee
We say that an \textbf{strong product assumption} is satisfied if instead of \eref{weakprodass}, we have
for every $n>0$ and $P, Q\in\Pi_n$, $PQ\in \Pi_{A^*n}$.
\end{definition}

\begin{uda}\label{uda:prodass}
{\rm
In the setting of Example~\ref{uda:hermite}, if $P,Q\in\Pi_n$, then $PQ=R\phi_{\bs 0}$ for some $R\in \Pi_{2n}$. So, the  product assumption holds trivially.
The strong product assumption does not hold.
However, if $P,Q\in\Pi_n$, then $PQ\in \mathsf{span}\{\phi_{\k,\sqrt{2}} : \lambda_k <n\sqrt{2}\}$.
The manifold case is discussed below in Remark~\ref{rem:steinerberger}.
\qed}
\end{uda}

\begin{rem}\label{rem:steinerberger}
{\rm
One of the referees of our paper has pointed out three recent references, \cite{steinerberger2017spectral, lu2019approximating, lu2018pointwise} on the subject of the product assumption. 
The first two of these deal with the manifold case (Example~\ref{uda:manifold}).
The paper \cite{lu2018pointwise} extends the results in \cite{lu2019approximating} to the case when the functions $\phi_k$ are eigenfunctions of a more general elliptic operator.
Since the  results in these two papers are similar qualitatively, we will comment on \cite{steinerberger2017spectral, lu2019approximating}.

In this remark only, let  $K_t(x,y)=\sum_k \exp(-\lambda_k^2t)\phi_k(x)\phi_k(y)$. 
Let $\lambda_k,\lambda_j< n$.
In \cite{steinerberger2017spectral}, Steinerberger relates $E_{An}(2,\phi_k\phi_j)$ (see \eqref{degapprox} below for definition) with 
$$\left\|\int_\XX K_t(\circ,y)(\phi_k(y)-\phi_k(\circ)) (\phi_j(y)-\phi_j(\circ))d\mu^*(y)\right\|_{2,\mu^*}.
$$
While this gives some insight into the product assumption, the results are inconclusive about the product assumption as stated.
Also, it is hard to verify whether the conditions mentioned in the paper are satisfied for a given manifold.

In \cite{lu2019approximating}, Lu, Sogge, and Steinerberger show that for any $\epsilon, \delta>0$, there exists a subspace $V$ of  dimension $\O_\delta(\epsilon^{-\delta}n^{1+\delta})$ such that for all $\phi_k,\phi_j\in\Pi_n$, $\inf_{P\in V}\|\phi_k\phi_j-P\|_{2,\mu^*}\le \epsilon$. 
The subspace $V$ does not have to be $\Pi_{An}$ for any $A$.
Since the dimension of $\mathsf{span}\{\phi_k\phi_j\}$ is $\O(n^2)$, the result is meaningful only if $0<\delta<1$ and $\epsilon\ge n^{1-1/\delta}$. 

In \cite[Theorem~6.1]{geller2011band}, Geller and Pesenson have shown that the strong product assumption (and hence, also the product assumption) holds in the manifold case when the manifold is a compact homogeneous manifold.
We have extended this theorem in \cite[Theorem~A.1]{modlpmz} for the case of eigenfunctions of general elliptic partial differential operators on arbitrary compact, smooth manifolds provided that the coefficient functions in the operator satisfy some technical conditions.
\qed}
\end{rem}
In our results in Section~\ref{bhag:mainresults}, we will need  the following condition, which serves the purpose of gradient in many of our earlier theorems on manifolds.

\begin{definition}\label{def:bernstein}
We say that the system $\Xi$ satisfies \textbf{Bernstein-Lipschitz condition} if for every $n>0$, there exists $B_n>0$ such that
\be\label{bernstein_der}
|P(x)-P(y)|\le B_n\rho(x,y)\|P\|_\infty, \qquad x, y\in\XX, \ P\in \Pi_n.
\ee
\end{definition}

\begin{rem}\label{rem:bernstein}
{\rm
Both in the manifold case and the Hermite case, $B_n=cn$ for some constant $c>0$. 
A proof in the Hermite case can be found in  \cite{mohapatrapap}, and in the manifold case in \cite{frankbern}.
\qed}
\end{rem}

\subsection{Smoothness classes}\label{bhag:smooth}
We define next the smoothness classes of interest here.
\begin{definition}\label{def:wtdef}
A function $w :\XX\to \RR$ will be called a \textbf{weight function} if  $w\phi_k\in C_0(\XX)\cap L^1(\XX)$ for all $k$.
If $w$ is a weight function,  we define
\be\label{degapprox}
E_n(w;p,f)=\min_{P\in \Pi_n}\|f-Pw\|_{p,\mu^*}, \qquad n>0,  1\le p \le \infty,\ f\in L^p(\XX).
\ee
We will omit the mention of $w$ if $w\equiv 1$ on $\XX$.
\end{definition}

We find it convenient to denote by $X^p$ the space $\{f\in L^p(\XX) : \lim_{n\to\infty}E_n(p,f)=0\}$; i.e., $X^p=L^p(\XX)$ if $1\le p<\infty$ and $X^\infty=C_0(\XX)$.

\begin{definition}\label{def:sobolev}
Let $1\le p\le \infty$, $\gamma>0$, and $w$ be a weight function. \\
{\rm (a)} For $f\in L^p(\XX)$, we define
\be\label{sobnorm}
\|f\|_{W_{\gamma,p,w}}=\|f\|_{p,\mu^*}+\sup_{n>0}n^\gamma E_n(w;p,f),
\ee
and note that
\be\label{sobnormuseful}
\|f\|_{W_{\gamma,p, w}}\sim \|f\|_{p,\mu^*}+\sup_{n\in\ZZ_+}2^{n\gamma}E_{2^n}(w;p,f).
\ee
The space $W_{\gamma,p,w}$ comprises all $f$ for which $\|f\|_{W_{\gamma,p,w}} <\infty$.\\
{\rm (b)}
We write $C^\infty_w=\bigcap_{\gamma>0}W_{\gamma,\infty,w}$.
If $B$ is a ball in $\XX$, $C^\infty_w(B)$ comprises functions in $f\in C^\infty_w$ which are supported on $B$.\\
{\rm (c)} If $x_0\in\XX$, the space $W_{\gamma,p,w}(x_0)$ comprises functions $f$ such that there exists $r>0$ with the property that for every $\phi\in C^\infty_w(\mathbb{B}(x_0,r))$, $\phi f\in W_{\gamma,p,w}$. 
\end{definition}

\begin{rem}\label{rem:smoothness}
{\rm
In both the manifold case and the Hermite case, characterizations of the smoothness classes $W_{\gamma,p}$ are available in terms of constructive properties of the functions, such as the number of derivatives, estimates on certain moduli of smoothness or $K$-functionals etc. 
In particular, the class $C^\infty$ coincides with the the class of infinitely differntiable functions vanishing at infinity.
\qed}
\end{rem}
We can now state another assumption which will be needed in studying local approximation.

\begin{definition}\label{def:partionunity}
(\textbf{Partition of unity})
For every $r>0$, there exists a  countable family $\mathcal{F}_r=\{\psi_{k,r}\}_{k=0}^\infty$ of functions in $C^\infty$ with the following properties:
\begin{enumerate}
\item Each $\psi_{k,r}\in \mathcal{F}_r$ is supported on $\mathbb{B}(x_k,r)$ for some $x_k\in\XX$.
\item For every $\psi_{k,r}\in\mathcal{F}_r$ and $x\in\XX$, $0\le \psi_{k,r}(x)\le 1$.
\item For every $x\in \XX$, there exists a finite subset $\mathcal{F}_r(x)\subseteq \mathcal{F}_r$ such that
\be\label{partitionsum}
\sum_{\psi_{k,r}\in \mathcal{F}_r(x)}\psi_{k,r}(y)=1, \qquad y\in\mathbb{B}(x,r).
\ee
 \end{enumerate}
\end{definition}
We note some obvious observations about the partition of unity without the simple proof.
\begin{prop}\label{prop:partitionofunity}
Let $r>0$, $\mathcal{F}_r$ be a partition of unity.\\
{\rm (a)}  Necessarily, $\disp\sum_{\psi_{k,r}\in \mathcal{F}_r(x)}\!\!\!\psi_{k,r}$ is supported on $\mathbb{B}(x, 3r)$.\\
{\rm (b)} For $x\in\XX$, $\sum_{\psi_{k,r}\in\mathcal{F}_r}\psi_{k,r}(x)=1$. 
\end{prop}

\noindent\textbf{The constant convention}
\emph{
In the sequel, $c, c_1,\cdots$ will denote generic positive constants depending only on the fixed quantities under discussion such as $\Xi$, $q$, $\kappa,\kappa_1,\kappa_2$, the various smoothness parameters and the filters to be introduced. 
Their value may be different at different occurrences, even within a single formula.
The notation $A\sim B$ means $c_1A \le B\le c_2A$.\qed
}

We end this section by defining a kernel which plays a central role in this theory.

Let $H : [0,\infty)\to \RR$ be a compactly supported function. In the sequel, we define 
\be\label{basickerndef}
\Phi_N(H;x,y)=\sum_{k=0}^\infty H(\lambda_k/N)\phi_k(x)\phi_k(y), \qquad N >0, \ x, y\in\XX.
\ee
If $S\ge 1$ is an integer, and $H$ is $S$ times continuously differentiable, we introduce the notation
$$
\||H|\|_S:=\max_{0\le k\le S}\max_{x\in\mathbb{R}}|H^{(k)}(x)|.
$$
The following proposition recalls an important property of these kernels. 
Proposition~\ref{prop:kernloc}  is proved in \cite{mauropap}, and more recently  in much greater generality in \cite[Theorem~4.3]{tauberian}.
\begin{prop}\label{prop:kernloc}
Let  $S>q$ be an integer, $H:\mathbb{R}\to \mathbb{R}$ be an even, $S$ times continuously differentiable, compactly supported function. 
 Then for every $x,y\in \mathbb{X}$, $N>0$,
\begin{equation}\label{kernlocest}
| \Phi_N(H;x,y)|\le \frac{cN^{q}\||H|\|_S}{\max(1, (N\rho(x,y))^S)}.
\end{equation}
\end{prop}
In the sequel, let $h:\RR\to [0,1]$ be a fixed, infinitely differentiable, even function, non-increasing on $[0,\infty)$, with $h(t)=1$ if $|t|\le 1/2$ and $h(t)=0$ if $t\ge 1$.
If $\nu$ is any measure having a bounded total variation on $\XX$, we define 
\be\label{gensummopdef}
\sigma_n(\nu, h;f)(x)=\int_\XX \Phi_n(h;x,y)f(y)d\nu(y).
\ee
We will omit the mention of $h$ in the notations; e.g., write $\Phi_n(x,y)=\Phi_n(h;x,y)$, and the mention of $\nu$ if $\nu=\mu^*$. In particular,
\be\label{summopdef}
\sigma_n(f)(x)=\sum_{k=0}^\infty h\left(\frac{\lambda_k}{n}\right)\hat{f}(k)\phi_k(x), \qquad n>0, \ x\in\XX, f\in L^1(\XX)+C_0(\XX),
\ee
where for $f\in L^1+C_0$, we write
\be\label{fourcoeffdef}
\hat{f}(k)=\int_\XX f(y)\phi_k(y)d\mu^*(y).
\ee

\subsection{Measures}\label{bhag:measures}
In this section, we describe the terminology involving measures.
\begin{definition}\label{absregularmeasuredef}
Let $d\ge 0$. A  measure $\nu\in \mathcal{M}$  will be called \textbf{$ d$--regular} if
\begin{equation}\label{regulardef}
|\nu|(\mathbb{B}(x,r))\le c(r+d)^q, \qquad x\in\mathbb{X}.
\end{equation}
The infimum of all constants $c$ which work in \eqref{regulardef} will be denoted by $|\!|\!|\nu|\!|\!|_{R,d}$, and the class of all  $d$--regular measures will be denoted by $\mathcal{R}_d$. 
\end{definition}

For example, $\mu^*$ itself is in  ${\cal R}_0$ with $|\!|\!|\mu^*|\!|\!|_{R,0}\le \kappa$ (cf. \eqref{ballmeasurecond}).
More generally, if $w\in C_0(\XX)$ then the measure $wd\mu^*$ is ${\cal R}_0$ with $|\!|\!|\mu^*|\!|\!|_{R,0}\le \kappa\|w\|_{\infty,\mu^*}$.

\begin{definition}\label{def:mzquadmeasure}
 {\rm (a)} A sequence $\{\nu_n\}$ of measures  on $\XX$ is called an \textbf{admissible quadrature measure sequence} if   the sequence $\{|\nu_n|(\XX)\}$ has polynomial growth and
\be\label{mzquadrature}
\int_\XX Pd\nu_n=\int_\XX Pd\mu^*, \qquad P\in \Pi_n, \ n\ge 1.
\ee
{\rm (b)} A sequence $\{\nu_n\}$ of measures  on $\XX$ is called an \textbf{admissible product quadrature measure sequence} if   the sequence $\{|\nu_n|(\XX)\}$ has polynomial growth and
\be\label{mz_prod_quadrature}
\int_\XX P_1P_2d\nu_n=\int_\XX P_1P_2d\mu^*, \qquad P_1, P_2\in \Pi_n, \ n\ge 1.
\ee
{\rm (c)} By abuse of terminology, we will say that a measure $\nu_n$ is an \textbf{admissible quadrature measure} (respectively, an \textbf{admissible product quadrature measure}) of order $n$ if $|\nu_n|\le c_1n^c$ (with constants independent of $n$) and \eref{mzquadrature} (respectively, \eref{mz_prod_quadrature}) holds.
\end{definition}

In the case when $\XX$ is compact, a well known theorem called Tchakaloff's theorem \cite[Exercise~2.5.8, p.~100]{rivlin1974chebyshev} shows the existence of   admissible product quadrature measures (even  finitely supported probability measures).
However, in order to construct such measures, it is much easier to prove the existence of admissible quadrature measures, as we will do in Theorem~\ref{theo:mztheo}, and then use one of the product assumptions to derive admissible product quadrature measures.

\begin{uda}\label{uda:manifoldquad}
{\rm
In the manifold case, let the strong product assumption hold as in Remark~\ref{rem:steinerberger}. 
If $n\ge 1$ and $\C\subset\XX$ is a finite subset satisfying the assumptions of  Theorem~\ref{theo:mztheo}, then the theorem asserts the existence of an admissible quadrature measure supported on $\C$. 
If $\{\nu_n\}$ is an admissible quadrature measure sequence, then $\{\nu_{A^*n}\}$ is an admissible product quadrature measure sequence.
In particular, there exist finitely supported admissible product quadrature measures of order $n$ for every $n\ge 1$. 
\qed}
\end{uda}
\begin{uda}\label{uda:hermitequad}
{\rm
We consider the Hermite case as in Example~\ref{uda:hermite}. For every $a>0$ and $n\ge 1$, Theorem~\ref{theo:mztheo} applied with the system $\Xi_a$ yields admissible quadrature measures of order $n$ supported on finite subsets of $\RR^q$ (in fact, of $[-cn,cn]^q$ for an appropriate $c$). In particular, an admissible quadrature measure of order $n\sqrt{2}$ for $\Xi_{\sqrt{2}}$ is an admissible product quadrature measure of order $n$ for $\Xi=\Xi_1$.
\qed}
\end{uda}

\subsection{Eignets}\label{bhag:eignet}

The notion of an eignet defined below is a generalization of the various kernels described in the examples in Section~\ref{bhag:motive}.

\begin{definition}\label{def:eignet}
A function $b :[0,\infty)\to (0,\infty)$ is called a \textbf{smooth mask} if $b$ is  non-increasing, and there exists $B^*=B^*(b)\ge 1$ such that the mapping $t\mapsto b(B^*t)/b(t)$ is fast decreasing. 
A function $G :\XX\times\XX\to \RR$ is called a \textbf{smooth kernel} if there exists a measurable function $W=W(G) :\XX\to\RR$ such that we have a formal expansion (with a smooth mask $b$)
\be\label{mercer}
W(y)G(x,y)=\sum_k b(\lambda_k)\phi_k(x)\phi_k(y), \qquad x, y\in \XX.
\ee
If $m\ge 1$ is an integer, an \textbf{eignet} with $m$ neurons is a function of the form $x\mapsto \sum_{k=1}^m a_k G(x,y_k)$ for $y_k\in\XX$.
\end{definition}

\begin{uda}\label{uda:eignetexample}
{\rm
In the manifold case, the notion of eignet includes all the examples stated in Section~\ref{bhag:motive}  with $W\equiv 1$, except for the example of smooth ReLU function described in Example~\ref{uda:sphere}. 
In the Hermite case,  \eref{gaussianexpansion} shows that  the kernel $\disp G(\x,\y)=\exp\left(-|\x-\frac{\sqrt{3}}{2}\y|_2^2\right)$  defined on $\RR^q\times\RR^q$ is a smooth kernel, with $\lambda_\k=|\k|_1$, $\phi_\k$ as in Example~\ref{uda:genhermite}, and $\disp b(t)=\left(\frac{3}{2\pi}\right)^{-q/2}3^{-t/2}$. 
The function $W$ here is $W(\y)=\exp(-|\y|_2^2/4)$. 
\qed}
\end{uda}

\begin{rem}\label{rem:eignet}
{\rm
It is possible to relax the conditions on the mask in Definition~\ref{def:eignet}. 
Firstly,  the condition that $b$ should be non-increasing is made only to simplify our proofs. It is not difficult to modify them without this assumption. 
Secondly, let $b_0 : [0,\infty)\to\RR$ satisfy $|b_0(t)|\le b_1(t)$ for a smooth mask $b_1$ as stipulated in that definition.
Then the function $b_2=b+2b_1$ is a smooth mask, and so is $b_1$. 
Let $G_j(x,y)=\sum_{k=0}^\infty b_j(\lambda_k)\phi_k(x)\phi_k(y)$, $j=0,1,2$. 
Then $G_0(x,y)=G_2(x,y)-2G_1(x,y)$.
Therefore, all of the results in Sections~\ref{bhag:mainresults} and \ref{bhag:pfsect} can be applied once with $G_2$ and once with $G_1$ to obtain a corresponding result for $G_0$, with different constants.
For this reason, we will simplify our presentation by assuming the apparently restrictive conditions stipulated in Definition~\ref{def:eignet}.
In particular, this includes the example of the smooth ReLU network described in Example~\ref{uda:sphere}. 
\qed}
\end{rem}
\begin{definition}\label{def:eignetkern}
Let $\nu$ be a measure on $\XX$ (signed or having bounded variation), and $G\in C_0(\XX\times\XX)$.
 We define
\be\label{phinder}
\mathcal{D}_{G,n}(x,y)=\sum_{k=0}^\infty h\left(\lambda_k/n\right)b(\lambda_k)^{-1}\phi_k(x)\phi_k(y),\qquad n\ge 1, \ x,y\in\XX,
\ee
and
\be\label{eignetkern}
\mathbb{G}_n(\nu;x,y)=\int_\XX G(x,z)W(z)\mathcal{D}_{G,n}(z,y)d\nu(z).
\ee
\end{definition}

\begin{rem}\label{rem:no_of_neurons}
{\rm
Typically, we will use an approximate product quadrature measure sequence in place of the measure $\nu$, where each of the measures in the sequence is finitely supported, to construct a sequence of networks. 
In the case when $\XX$ is compact, Tchakaloff's theorem shows that there exists an approximate product quadrature measure of order $m$ supported on $(\mathsf{dim}(\Pi_m)+1)^2$ points.
Using this measure in place of $\nu$, one obtains a pre-fabricated eignet $\mathbb{G}_n(\nu)$ with $(\mathsf{dim}(\Pi_m)+1)^2$ neurons.
However, this is not an actual construction.
In  the presence of the product assumption, Theorem~\ref{theo:mztheo} leads to the pre-fabricated networks $\mathbb{G}_n$ in a constructive manner with the number of neurons as stipulated in that theorem.
\qed}
\end{rem}
\bhag{Main results}\label{bhag:mainresults}
In this section, we assume the Bernstein-Lipschitz condition (Definition~\ref{def:bernstein})  in all the theorems.
We note that the measure $\mu^*$ may not be a probability measure. 
Therefore, we take the help of an auxiliary function $f_0$ to define a probability measure  as follows.
Let $f_0\in C_0(\XX)$, $f_0\ge 0$ for all $x\in \XX$, and $d\nu^*=f_0d\mu^*$ be a probability measure.
Necessarily, $\nu^*$ is $0$-regular, and $\tn\nu^*\tn_{R,0} \le \|f_0\|_{\infty,\mu^*}$.
We assume noisy data of the form $(y,\epsilon)$, with a joint probability distribution $\tau$ defined for Borel subsets of $\XX\times\Omega$ for some measure space $\Omega$, and with $\nu^*$ being the marginal distribution of $y$ with respect to $\tau$. 
Let $\mathcal{F}(y,\epsilon)$ be a random variable following the law $\tau$, and denote
\be\label{Fdef} 
f(y)=\mathbb{E}_\tau(\mathcal{F}(y,\epsilon)|y).
\ee
 It is easy to verify using Fubini's theorem that if $\mathcal{F}$ is integrable with respect to $\tau$ then
for any $x\in\XX$,
\be\label{iterexp}
\mathbb{E}_\tau(\mathcal{F}(y,\epsilon)\Phi_n(x,y))=\sigma_n(\nu^*;f)(x):=\int_\XX f(y)\Phi_n(x,y)d\nu^*(y).
\ee

Let  $Y$ be a random sample from $\tau$, and $\{\nu_n\}$ be an admissible product quadrature sequence in the sense of Definition~\ref{def:mzquadmeasure}. We define (cf. \eref{eignetkern})
\be\label{sampleeignetdef}
\mathcal{G}_n(Y; \mathcal{F})(x)=\mathcal{G}_n(\nu_{B^*n},Y; \mathcal{F})(x)=\frac{1}{|Y|}\sum_{(y,\epsilon)\in Y}\mathcal{F}(y,\epsilon)\mathbb{G}_n(\nu_{B^*n};x,y), \qquad x\in\XX, \ n=1,2,\cdots,
\ee
where $B^*$ is as in Definition~\ref{def:eignet}.
\begin{rem}\label{rem:no_of_neurons_bis}
{\rm
We note that the networks $\mathbb{G}_n$ are pre-fabricated, independently of the data. Therefore, effectively, the network $\mathcal{G}_n$ has only $|Y|$ terms depending upon the data.
\qed}
\end{rem}

Our first theorem describes local function recovery using local sampling.
We may interpret it in the spirit of distributed learning as in \cite{chui2019realization,lin2019distributed}, where we are taking a linear combination of pre-fabricated networks $\mathbb{G}_n$ using the function values themselves as the coefficients.
The networks $\mathbb{G}_n$ have  essentially the same localization property as the kernels $\Phi_n$ (cf. Theorem~\ref{theo:eignetkern}).

\begin{theorem}\label{theo:distributed}
Let $x_0\in\XX$, and $r>0$. We assume the partition of unity, and find a function $\psi\in C^\infty$ supported on $\mathbb{B}(x_0,3r)$ which is equal to $1$ on $\mathbb{B}(x_0,r)$, $\mathfrak{m}=\int_\XX \psi d\mu^*$, and let $f_0=\psi/\mathfrak{m}$, $d\nu^*=f_0d\mu^*$.  We assume the rest of the set-up as described. 
If $f_0f\in W_{\gamma, \infty}$,
then for $0<\delta<1$, and $|Y|\ge cn^{q+2\gamma}r^q\log(nB_n/\delta)$,
\be\label{distributed}
\mathsf{Prob}_{\tau}\left(\left\{\left\|\frac{\mathfrak{m}}{|Y|}\sum_{(y,\epsilon)\in Y}\mathcal{F}(y,\epsilon)\mathbb{G}_n(\nu_{B^*n};\circ,y)-f\right\|_{\infty,\mu^*,\mathbb{B}(x_0,r)} \ge c_3n^{-\gamma}\right\}\right)\le \delta.
\ee
\end{theorem}

\begin{rem}\label{rem:subsample}
{\rm
If $\{y_1,\cdots,y_M\}$ is a random sample from some probability measure supported on $\XX$, $s=\sum_{\ell=1}^M f_0(y_\ell)$, and we construct a sub-sample using the distribution that associates the mass $f_0(y_j)/s$ with each $y_j$, then the probability of selecting points outside of the support of $f_0$ is $0$. This leads to a sub-sample $Y$. If $M\ge cn^{q+2\gamma}\log(nB_n/\delta)$, then  the Chernoff bound, Proposition~\ref{prop:concentration}(b), can be used to show that $|Y|$ is large as stipulated in Theorem~\ref{theo:distributed}.
\qed}
\end{rem}

Next, we state two inverse theorems.
Our first theorem obtains accuracy on  the estimation of the density $f_0$ using eignets instead of positive kernels.

\begin{theorem}\label{theo:densityest}
With the set-up as in Theorem~\ref{theo:eignetprob}, let $\gamma>0$, $f_0\in W_{\gamma,\infty}$, and 
$$
|Y|\ge \|f_0\|_{\infty,\mu^*}n^{q+2\gamma}\log\left(\frac{nB_n}{\delta}\right).
$$ 
Then, with $\mathcal{F}\equiv 1$,
\be\label{eignetprobdensity}
\mathsf{Prob}_{\tau}\left(\left\{\left\|\frac{1}{|Y|}\sum_{(y,\epsilon)\in Y}\mathbb{G}_n(\nu_{B^*n};\circ,y)-f_0\right\|_\infty \ge c_3n^{-\gamma}\right\}\right)\le \delta.
\ee
\end{theorem}

\begin{rem}\label{rem:nolimit}
{\rm
Unlike density estimation using positive kernels, there is no inherent limit on the accuracy predicted by \eref{eignetprobdensity} on the estimation of $f_0$. 
\qed}
\end{rem}

The following theorem gives a complete characterization of the local smoothness classes using eignets. 
In particular, Part (b) of the following theorem gives a solution to the inverse problem of determining what smoothness class the target function belongs to near each point of $\XX$.
In theory, this leads to a \textbf{data-based detection} of singularities and sparsity analogous to what is assumed in 
\cite{chui2019realization}, but in much more general setting.
\begin{theorem}\label{theo:loceignet}
Let $f_0\in C_0(\XX)$, $f_0(x)\ge 0$ for all $x\in \XX$, and $d\nu^*=f_0d\mu^*$ be a probability measure, $\tau$, $\mathcal{F}$, and $f$ be as described above. 
We assume the partition of unity and the product assumption.
Let $S\ge q+2$, $0<\gamma\le S$, $x_0\in \XX$, $0<\delta<1$.
For each $j\ge 0$, suppose that $Y_j$ is a random sample from $\tau$ with $|Y_j|\ge  2c_12^{j(q+2S)}\tn\nu^*\tn_{R,0}\log(c2^{2j}B_{2^j}/\delta)$. 
Then with $\tau$-probability $\ge 1-\delta$,\\
{\rm (a)} If $f_0f\in W_{\gamma,\infty}(x_0)$ then there exists a ball $\BB$ centered at $\x_0$ such that 
\be\label{eignettauest}
\sup_{j\ge 1}2^{j\gamma}\|\mathcal{G}_{2^j}(Y_j; \mathcal{F})-\mathcal{G}_{2^{j-1}}(Y_j; \mathcal{F})\|_{\infty,\mu^*,\BB} <\infty.
\ee
{\rm (b)} If there exists a ball $\BB$ centered at $x_0$ for which \eref{eignettauest} holds, then $f_0f\in W_{\gamma,\infty,\phi_0}(x_0)$.
\end{theorem}

\bhag{Preparatory results}\label{bhag:ddrprep}

We prove a lower bound on $\mu^*(\mathbb{B}(x,r))$ for $x\in \XX$ and $0<r\le 1$ (cf. \cite{grigorlyan2heat}).

\begin{prop}\label{prop:ballmeasurelow}
We have
\be\label{ballmeasurelow}
\mu^*(\mathbb{B}(x,r))\ge cr^q, \qquad 0<r\le 1, \ x\in\XX.
\ee
\end{prop}

In order to prove the proposition, we recall a lemma, proved in \cite[Proposition~5.1]{eignet}.
\begin{lemma}\label{lemma:critical}
Let $\nu\in {\cal R}_d$,  $N>0$. If $g_1:[0,\infty)\to [0,\infty)$ is a nonincreasing function, then for any $N>0$, $r>0$, $x\in\mathbb{X}$,
\begin{equation}\label{g1ineq}
N^q\int_{\Delta(x,r)}g_1(N\rho(x,y))d|\nu|(y)\le c\frac{2^{q}(1+(d/r)^q)q}{1-2^{-q}}|\!|\!|\nu|\!|\!|_{R,d}\int_{rN/2}^\infty g_1(u)u^{q-1}du.
\end{equation}
\end{lemma}

\noindent\textsc{Proof of Proposition~\ref{prop:ballmeasurelow}.}\\

Let $x\in\XX$, $r>0$ be fixed in this proof, although the constants will not depend upon these. 
In this proof, we write 
$$
K_t(x,y)=\sum_{k=0}^\infty \exp(-\lambda_k^2t)\phi_k(x)\phi_k(y).
$$
The Gaussian upper bound \eref{gaussianbd} shows that for $t>0$,
\be\label{pf1eqn1}
\int_{\Delta(x, r)}|K_t(x,y)|d\mu^*(y) \le \kappa_1 t^{-q/2}\int_{\Delta(x, r)} \exp(-\kappa_2\rho(x,y)^2/t)d\mu^*(y).
\ee
Using Lemma~\ref{lemma:critical}  with $d=0$, $d\nu=d\mu^*$, $g_1(u)=\exp(- u^2)$, $N=\sqrt{\kappa_2/t}$, we obtain for $r^2/t\ge (q-2)/\kappa_2$:
\bea\label{pf1eqn2}
\int_{\Delta(x, r)}|K_t(x,y)|d\mu^*(y) &\le& c\int_{Nr/2}^\infty u^{q-1}\exp(-u^2)du = c_1\int_{(Nr/2)^2}^\infty\!\!\! u^{q/2-1}e^{-u}du\nonumber\\
& \le& c_2(r^2/t)^{(q-2)/2}\exp(-\kappa_2r^2/(4t)).
\eea
Therefore, denoting in this proof only, $\kappa_0=\|\phi_0\|_\infty$, we obtain that
\be\label{pf1eqn3}
1=\int_\XX K_t(x,y)\phi_0(y)d\mu^*(y)\le \kappa_0\int_\XX |K_t(x,y)|d\mu^*(y) \le \kappa_0\kappa_2t^{-q/2}\mu^*(\mathbb{B}(x,r)) + c_3(r^2/t)^{(q-2)/2}\exp(-\kappa_2r^2/(4t)).
\ee
We now choose $t\sim r^2$ so that $c_3(r^2/t)^{(q-2)/2}\exp(-\kappa_3r^2/(4t)) \le 1/2$ to obtain \eref{ballmeasurelow} for $r\le c_4$. 
The estimate is clear for $c_4<r\le 1$.
\qed

Next, we prove some results about the system $\{\phi_k\}$.

\begin{lemma}\label{lemma:christoffel}
For $n\ge 1$, we have
\be\label{christoffel}
\sum_{\lambda_k<n}\phi_k(x)^2 \le cn^q, \qquad x\in\XX.
\ee
and
\be\label{dimension}
\mathsf{dim}(\Pi_n)\le cn^q\mu^*(\mathbb{K}_n).
\ee
In particular, the function $n\mapsto \mathsf{dim}(\Pi_n)$ has polynomial growth.
\end{lemma}

\begin{Proof}\ % of Lemma~\ref{lemma:christoffel}
The Gaussian upper bound with $x=y$ implies that 
$$
\sum_{k=0}^\infty \exp(-\lambda_k^2t)\phi_k(x)^2\le ct^{-q/2}, \qquad 0<t\le 1, \ x\in\XX.
$$
The estimate \eref{christoffel} follows from a Tauberian theorem \cite[Proposition~4.1]{frankbern}.
The essential compactness now shows that for any $R>0$,
$$
\int_{\XX\setminus \mathbb{K}_n}\sum_{\lambda_k<n}\phi_k(x)^2d\mu^*(x)\le \left\{\sup_{x\in\XX\setminus\mathbb{K}_n}\sum_{\lambda_k<n}\phi_k(x)^2\right\}^{1/2}\int_{\XX\setminus\mathbb{K}_n}\left(\sum_{\lambda_k<n}\phi_k(x)^2\right)^{1/2}d\mu^*(x) \le cn^{-R}.
$$
In particular,
$$
\mathsf{dim}(\Pi_n)=\int_\XX \sum_{\lambda_k<n}\phi_k(x)^2d\mu^*(x)\le \int_{\mathbb{K}_n}\sum_{\lambda_k<n}\phi_k(x)^2d\mu^*(x) +cn^{-R}\le cn^q\mu^*(\mathbb{K}_n).
$$
\end{Proof}

Next, we prove some properties of the operators $\sigma_n$ and diffusion polynomials.
The following proposition follows easily from Lemma~\ref{lemma:critical} and Proposition~\ref{prop:kernloc}. (cf. \cite{sloanfest, eignet}).

\begin{prop}\label{prop:critical}
Let  $S$, $H$ be as in Proposition~\ref{prop:kernloc}, $d>0$, $\nu\in\mathcal{R}_d$, and  $x\in\mathbb{X}$. \\
{\rm (a)} If  $r\ge  1/N$, then
\begin{equation}\label{phiintaway}
\int_{\Delta(x,r)}|\Phi_N(H;x,y)|d|\nu|(y) \le c(1+(dN)^q)(rN)^{-S+q}|\!|\!|\nu|\!|\!|_{R,d}\||H|\|_S.
\end{equation}
{\rm (b)} We have
\begin{equation}\label{phiinttotal}
\int_\mathbb{X}|\Phi_N(H;x,y)|d|\nu|(y)\le c(1+(dN)^q)|\!|\!|\nu|\!|\!|_{R,d}\||H|\|_S,
\end{equation} 
\begin{equation}\label{philpnorm}
\|\Phi_N(H;x,\circ)\|_{\nu;\mathbb{X},p} \le cN^{q/p'}(1+(dN)^q)^{1/p}|\!|\!|\nu|\!|\!|_{R,d}^{1/p}\||H|\|_S,
\end{equation}
and
\begin{equation}\label{phiintnorm}
\left\|\int_\mathbb{X}|\Phi_N(H;\circ,y)|d|\nu|(y)\right\|_p \le c(1+(dN)^q)^{1/p'}|\!|\!|\nu|\!|\!|_{R,d}^{1/p'}(|\nu|(\mathbb{X}))^{1/p}\||H|\|_S.
\end{equation}
\end{prop}

The following lemma is well known; a proof is given in \cite[Lemma~5.3]{heatkernframe}.

\begin{lemma}\label{lemma:young}
Let $(\Omega_1,\nu)$, $(\Omega_2,\tau)$ be sigma--finite measure spaces, $\Psi : \Omega_1\times\Omega_2 \to \RR$ be $\nu\times\tau$--integrable,
\be\label{gennormdef}
M_\infty:=\nu\!\!-\!\!\esssup_{x\in\Omega_1}\int_{\Omega_2}|\Psi(x,y)|d\tau(y)<\infty, \quad M_1:=\tau\!\!-\!\!\esssup_{y\in\Omega_2}\int_{\Omega_1}|\Psi(x,y)|d\nu(x)<\infty,
\ee
and formally, for $\tau$--measurable functions $f :\Omega_2\to\RR$,
$$
T(f,x):=\int_{\Omega_2} f(y)\Psi(x,y)d\tau(y), \qquad x\in\Omega_1.
$$
Let $1\le p\le \infty$. If $f\in L^p(\tau;\Omega_2)$ then $T(f,x)$ is defined for $\nu$--almost all $x\in\Omega_1$, and 
\be\label{genopbd}
\|Tf\|_{\nu;\Omega_1,p}\le  M_1^{1/p}M_{\infty}^{1/p'}\|f\|_{\tau;\Omega_2,p}, \qquad f\in L^p(\Omega_2,\tau).
\ee
\end{lemma}
\begin{theorem}\label{theo:goodapprox}
Let $n>0$. If $P\in\Pi_{n/2}$, then $\sigma_n(P)=P$. Also,
for any $p$ with $1\le p\le\infty$, 
\be\label{opbd}
\|\sigma_n(f)\|_p \le c\|f\|_p, \qquad f\in L^p.
\ee 
If $1\le p\le \infty$, and $f\in L^p(\XX)$, then
\be\label{goodapprox}
E_n(p,f)\le \|f-\sigma_n(f)\|_{p,\mu^*}\le cE_{n/2}(p,f).
\ee
\end{theorem}
\begin{Proof}\ % of Theorem~\ref{theo:goodapprox}
The fact that $\sigma_n(P)=P$ for all $P\in\Pi_{n/2}$ is verified easily using the fact that $h(t)=1$ for $0\le t\le 1/2$. 
Using \eref{phiinttotal} with $\mu^*$ in place of $|\nu|$ and $0$ in place of $d$, we see that
$$
\sup_{x\in\XX}\int_\XX|\Phi_n(x,y)|d\mu^*(y) \le c.
$$
The estimate \eref{opbd} follows using Lemma~\ref{lemma:young}. The estimate \eref{goodapprox} is now routine to prove.
\end{Proof}

\begin{prop}\label{prop:rangeineq}
For $n\ge 1$, $P\in \Pi_n$, $1\le p\le \infty$, and $S>0$, we have
\be\label{rangeineq}
\|P\|_{p,\mu^*, \XX\setminus\mathbb{K}_{2n}} \le c(S)n^{-S}\|P\|_{p,\mu^*,\XX}.
\ee
\end{prop}

\begin{Proof}\ % of Proposition~\ref{prop:rangeineq}
In this proof, all constants will depend upon $S$. Using Schwarz inequality and essential compactness, it is easy to deduce that
\be\label{pf2eqn1}
\sup_{x\in \XX\setminus \mathbb{K}_{2n}}\int_\XX |\Phi_{2n}(x,y)|d\mu^*(y) \le c_1n^{-S},\quad \sup_{y\in\XX}\int_{\XX\setminus \mathbb{K}_{2n}}|\Phi_{2n}(x,y)|d\mu^*(x) \le c_1n^{-S}.
\ee
Therefore, a use of Lemma~\ref{lemma:young} shows that
$$
\|\sigma_{2n}(f)\|_{p,\mu^*,\XX\setminus \mathbb{K}_{2n}}\le cn^{-S}\|f\|_p.
$$
We use $P$ in place of $f$ to obtain \eref{rangeineq}.
\end{Proof}

\begin{prop}\label{prop:nikolskii}
Let $n\ge 1$, $P\in \Pi_n$, $0<p<r\le \infty$. Then 
\be\label{nikolskii}
\|P\|_r\le cn^{q(1/p-1/r)}\|P\|_p, \qquad \|P\|_p\le c\mu^*(\mathbb{K}_{2n})^{1/p-1/r}\|P\|_r.
\ee
\end{prop}
\begin{Proof}\ %of Proposition~\ref{prop:nikolskii}
The first part of \eref{nikolskii} is proved in \cite[Lemma~5.4]{heatkernframe}. 
In that paper, the measure $\mu^*$ is assumed to be a probability measure, but this assumption was not used in this proof.
The second estimate follows easily from Proposition~\ref{prop:rangeineq}.
\end{Proof}

\begin{lemma}\label{lemma:polyproduct}
Let $R, n>0$, $P_1,P_2\in \Pi_n$, $1\le p, r, s\le \infty$. If the product assumption holds, then
\be\label{polyproduct}
E_{A^*n}(\phi_0; p, P_1P_2)\le cn^{-R}\|P_1\|_r\|P_2\|_s.
\ee
\end{lemma}

\begin{Proof} \ %of Lemma~\ref{lemma:polyproduct}
In view of essential compactness, Proposition~\ref{prop:nikolskii} implies that for any $P\in \Pi_n$, $1\le r\le\infty$, $\|P\|_2\le c_1n^c\|P\|_r$. Therefore, using Schwarz inequality, Parseval identity, and Lemma~\ref{lemma:christoffel}, we conclude that
\be\label{pf3eqn1}
\sum_k |\hat{P}(k)| \le (\mathsf{dim}(\Pi_n))^{1/2}\|P\|_2\le c_1n^c\|P\|_r.
\ee
Now, the product assumption implies that for $p=1,\infty$, and $\lambda_k, \lambda_j <n$, there exists $R_{j,k,n}\in\Pi_{A^*n}$ such that for any $R>0$,
\be\label{pf3eqn2}
\|\phi_k\phi_j-R_{j,k,n}\phi_0\|_p \le cn^{-R-2c},
\ee
where $c$ is the constant appearing in \eref{pf3eqn1}.
The convexity inequality
$$
\|f\|_p\le \|f\|_\infty^{1/p'}\|f\|_1^{1/p}
$$
shows that \eref{pf3eqn2} is valid for all $p$, $1\le p\le \infty$.
So, using \eref{pf3eqn1}, we conclude that
$$
\left\|P_1P_2-\sum_{k,j}\widehat{P_1}(k)\widehat{P_2}(k)R_{j,k,n}\phi_0\right\|_p \le cn^{-R-2c}\left(\sum_k |\widehat{P_1}(k)|\right)\left(\sum_k |\widehat{P_2}(k)|\right)\le cn^{-R}\|P_1\|_r\|P_2\|_s.
$$
\end{Proof}

\bhag{Local approximation by diffusion polynomials}\label{bhag:locapprox}
In the sequel, we write $g(t)=h(t)-h(2t)$, and
\be\label{taudef}
\tau_j(f)=\begin{cases}
\sigma_1(f), & \mbox{ if $j=0$},\\
\sigma_{2^j}(f)-\sigma_{2^{j-1}}(f), &\mbox{ if $j=1,2,\cdots$.}
\end{cases}
\ee
We note that
\be\label{taukerndef}
\tau_j(f)(x)=\sigma_{2^j}(\mu^*,g;f)(x)=\int_\XX f(y)\Phi_{2^j}(g;x,y)d\mu^*(y), \qquad j=1,2,\cdots.
\ee
It is clear from Theorem~\ref{theo:goodapprox} that for any $p$, $1\le p\le \infty$,
\be\label{paleywiener}
f=\sum_{j=0}^\infty \tau_j(f), \qquad f\in X^p, 
\ee
with convergence in the sense of $L^p$.

\begin{theorem}\label{theo:paleywiener}
Let $1\le p\le \infty$, $\gamma>0$, $f\in X^p$, $x_0\in\XX$. We assume the partition of unity and the product assumption. \\
{\rm (a)} If $\BB$ is a ball centered at $x_0$, then 
\be\label{ballequiv}
\sup_{n\ge 0}2^{n\gamma}\|f-\sigma_{2^n}(f)\|_{p,\mu^*,\BB}\sim \sup_{j\ge 0}2^{j\gamma}\|\tau_j(f)\|_{p,\mu^*,\BB}.
\ee
{\rm (b)} If there exists a ball $\BB$ centered at $x_0$ such that
\be\label{op_implies_sobol}
\sup_{n\ge 0}2^{n\gamma}\|f-\sigma_{2^n}(f)\|_{p,\mu^*,\BB} \sim \sup_{j\ge 0}2^{j\gamma}\|\tau_j(f)\|_{p,\mu^*,\BB}<\infty,
\ee
then $f\in W_{\gamma,p,\phi_0}(x_0)$.\\
{\rm (c)} If $f\in W_{\gamma,p}(x_0)$, then there exists a ball $\BB$ centered at $x_0$ such that \eref{op_implies_sobol} holds.
\end{theorem}

\begin{rem}\label{rem:paleywiener}
{\rm
In the manifold case  (Example~\ref{uda:manifold}), $\phi_0\equiv 1$. 
So, the statements (b) and (c) in Theorem~\ref{theo:paleywiener} provide necessary and sufficient conditions for $f\in W_{\gamma,p}(x_0)$ in terms of the local rate of convergence of the globally defined operator $\sigma_n(f)$, respectively, the growth of the local norms of the operators $\tau_j$. 
In the Hermite case (Example~\ref{uda:hermite}), it is shown in \cite{tenswt} that  $f\in W_{\gamma,p,\phi_0}$ if and only if $f\in W_{\gamma,p}$.
Therefore, the statements (b) and (c) in Theorem~\ref{theo:paleywiener} provide similar necessary and sufficient conditions for $f\in W_{\gamma,p}(x_0)$ in this case as well.
\qed}
\end{rem}
The proof of Theorem~\ref{theo:paleywiener} is routine, but we sketch a proof for the sake of completeness.\\

\noindent\textsc{Proof of Theorem~\ref{theo:paleywiener}}\\

Part (a) is easy to prove using the definitions. \\
In the rest of this proof, we fix $S>\gamma+q+2$.
To prove part (b), let $\phi\in C^\infty$ be supported on $\BB$. Then there exists $\{R_n\in\Pi_{2^n}\}_{n=0}^\infty$ such that 
\be\label{pf4eqn1}
\|\phi-R_n\|_\infty\le c(\phi)2^{-nS}.
\ee
Further, Lemma~\ref{lemma:polyproduct} yields a sequence $\{Q_n\in\Pi_{A^*2^n}\}$ such that
\be\label{pf4eqn2}
\|R_n\sigma_{2^n}(f)-\phi_0Q_n\|_p \le c2^{-nS}\|R_n\|_\infty\|\sigma_{2^n}(f)\|_p \le c(\phi)2^{-nS}\|f\|_p.
\ee
Hence,
\begin{eqnarray*}
E_{A^*2^n}(\phi_0;p,f\phi)&\le& \|f\phi-\phi_0Q_n\|_p\le c(\phi)2^{-nS}\|f\|_p +\|f\phi-\sigma_{2^n}(f)R_n\|_p\\
&\le& c(\phi)2^{-nS}\|f\|_p +\|(f-\sigma_{2^n}(f))\phi\|_p+\|\sigma_{2^n}(f)(\phi-R_n)\|_p \\
&\le& c(\phi)\left\{2^{-nS}\|f\|_p+\|f-\sigma_{2^n}(f)\|_{p,\mu^*,\mathbb{B}}+\|\sigma_{2^n}(f)\|_p\|\phi-R_n\|_\infty\right\}\\
&\le& c(\phi)2^{-nS}\|f\|_p+ c(\phi,f)(A^*2^{-n})^\gamma.
\end{eqnarray*}
Thus, $f\phi\in W_{\gamma,p,\phi_0}$ for every $\phi\in C^\infty$ supported on $\mathbb{B}$, and part (b) is proved.\\

To prove part (c), we observe that there exists $r>0$ such that for any $\phi\in C^\infty(\mathbb{B}(x_0,6r))$, $f\phi\in W_{\gamma,p}$. 
Using partition of unity (cf. Proposition~\ref{prop:partitionofunity}(a)), we find $\psi\in C^\infty(\mathbb{B}(x_0,6r))$ such that $\psi(x)=1$ for all $x\in\mathbb{B}(x_0,2r)$, and let $\mathbb{B}=\mathbb{B}(x_0,r)$.
In view of Proposition~\ref{prop:kernloc}, $|\Phi_{2^n}(x,y)|\le c(r)2^{-n(S-q)}$ for all $x\in \mathbb{B}$ and $y\in \XX\setminus \mathbb{B}(x_0,2r)$. 
Hence,
\bea\label{pf4eqn3}
\|\sigma_{2^n}((1-\psi)f)\|_p &\le& \left|\int_\XX |(1-\psi(y))f(y)\Phi_{2^n}(\circ,y)|d\mu^*(y)\right\|_p\nonumber\\
&=& \left|\int_{\XX\setminus \mathbb{B}(x_0,2r)} |(1-\psi(y))f(y)\Phi_{2^n}(\circ,y)|d\mu^*(y)\right\|_p\le  c(\psi,r)2^{-n(S-q)}\|f\|_p.
\eea
Recalling that $\psi(x)=1$ for $x\in\mathbb{B}$ and $S-q\ge \gamma+2$, we deduce that
\begin{eqnarray*}
\|f-\sigma_{2^n}(f)\|_{p,\mu^*,\mathbb{B}}&=&\|\psi f -\sigma_{2^n}(f)\|_{p,\mu^*,\mathbb{B}}\le \|\psi f-\sigma_{2^n}(\psi f)\|_{p,\mu^*,\mathbb{B}}+\|\sigma_{2^n}((1-\psi)f)\|_p\\
&\le& cE_{2^n}(\psi f)+c(\psi,r)2^{-n(S-q)}\|f\|_p \le c(r,\psi,f)2^{-n\gamma}.
\end{eqnarray*}
This proves part (c). \qed

Let $\{\Psi_n :\XX\times\XX\to \XX\}$ be a family of kernels (not necessarily symmetric). With a slight abuse of notation, we define when possible, for any measure $\nu$ with  bounded total variation on $\XX$,
\be\label{networdopdef}
\sigma(\nu,\Psi_n;f)(x)=\int_\XX f(y)\Psi_n(x,y)d\nu(y), \qquad x\in\XX,\ f\in L^1(\XX)+C_0(\XX),
\ee
and
\be\label{networktaudef}
\tau_j(\nu, \{\Psi_n\};f)=\begin{cases}
\sigma(\nu, \Psi_1;f), &\mbox{if $j=0$},\\
\sigma(\nu,\Psi_{2^j};f)-\sigma(\nu,\Psi_{2^{j-1}};f), &\mbox{ if $j=1,2,\cdots$.}
\end{cases}
\ee
As usual, we will omit the mention of $\nu$ when $\nu=\mu^*$.
\begin{cor}\label{cor:networkcor}
Let the assumptions of Theorem~\ref{theo:paleywiener} hold, and $\{\Psi_n :\XX\times\XX\to \XX\}$ be a seqence of kernels (not necessarily symmetric) with the property that both of the following functions of $n$ are fast decreasing.
\be\label{kernel_net_approx}
\sup_{x\in\XX}\int_\XX |\Psi_n(x,y)-\Phi_n(x,y)|d\mu^*(y), \quad \sup_{y\in\XX}\int_\XX |\Psi_n(x,y)-\Phi_n(x,y)|d\mu^*(x). 
\ee 
{\rm (a)} If $\BB$ is a ball centered at $x_0$, then 
\be\label{gen_ballequiv}
\sup_{n\ge 0}2^{n\gamma}\|f-\sigma(\Psi_{2^n};f)\|_{p,\mu^*,\BB}\sim \sup_{j\ge 0}2^{j\gamma}\|\tau_j(\{\Psi_n\};f)\|_{p,\mu^*,\BB}.
\ee
{\rm (b)} If there exists a ball $\BB$ centered at $x_0$ such that
\be\label{gen_op_implies_sobol}
\sup_{n\ge 0}2^{n\gamma}\|f-\sigma(\Psi_{2^n};f)\|_{p,\mu^*,\BB} \sim \sup_{j\ge 0}2^{j\gamma}\|\tau_j(\{\Psi_n\};f)\|_{p,\mu^*,\BB}<\infty,
\ee
then $f\in W_{\gamma,p,\phi_0}(x_0)$.\\
{\rm (c)} If $f\in W_{\gamma,p}(x_0)$, then there exists a ball $\BB$ centered at $x_0$ such that \eref{gen_op_implies_sobol} holds.
\end{cor}

\begin{Proof}\ % of Corollary~\ref{cor:networkcor}
In view of Lemma~\ref{lemma:young}, the assumption about the functions in \eref{kernel_net_approx} implies that $\|\sigma(\Psi_n;f)-\sigma_n(f)\|_p$ is fast decreasing.
\end{Proof}

\bhag{Quadrature formula}\label{bhag:quadrature}

The purpose of this section is to prove the existence of admissible quadrature measures in the general set-up as in this paper. 
The ideas are mostly developed already in our earlier works \cite{indiapap, mnw1, frankbern, modlpmz, mohapatrapap, chuigaussian}, but always requiring an estimate on the gradient of diffusion polynomials. 
Here, we use the Bernstein-Lipschitz condition (Definition~\ref{def:bernstein}) instead.

If $\C\subset K\subset\XX$,  we denote
\be\label{meshnorm}
\delta(K,\C)=\sup_{x\in K}\inf_{y\in\C}\rho(x,y), \qquad \eta(\C)=\inf_{x,y\in\C, x\not=y}\rho(x,y).
\ee
If $K$ is compact, $\epsilon>0$, a subset $\C\subset K$ is $\epsilon$-distinguishable if $\rho(x,y)\ge \epsilon$ for every $x, y\in \C$, $x\not=y$. The cardinality the maximal $\epsilon$-distinguishable subset of $K$ will be denoted by $H_\epsilon(K)$.

\begin{rem}\label{rem:uniformity}
{\rm
If $\C_1\subset\C$ is a maximal $\delta(K,\C)$-distinguishable subset of $\C$, $x\not=y$, then it is easy to deduce that
$$
\delta(K,\C)\le \eta(\C_1)\le 2\delta(K,\C),\qquad \delta(K,\C)\le \delta(K,\C_1)\le 2\delta(K,\C).
$$
In particular, by replacing $\C$ by $\C_1$, we can always assume that
\be\label{uniformity}
(1/2)\delta(K, \C)\le \eta(\C)\le 2\delta(K,\C).
\ee 
}
\end{rem}

\begin{theorem}\label{theo:mztheo}
We assume the Bernstein-Lipschitz condition. 
Let $n>0$, $\C_1=\{z_1,\cdots, z_M\}\subset \mathbb{K}_{2n}$ be a finite subset, $\epsilon>0$.\\
{\rm (a)}  There exists a constant $c(\epsilon)$ with the following property: if $\delta(\mathbb{K}_{2n},\C_1) \le c(\epsilon)\min(1/n, 1/B_{2n})$ then there exist non-negative numbers $W_k$ with
\be\label{mzwtest}
W_k\le c\delta(\mathbb{K}_{2n},\C_1)^q, \qquad \sum_{k=1}^M W_k\le c\mu^*(\mathbb{B}(\mathbb{K}_{2n},4\delta(\mathbb{K}_{2n},\C_1))).
\ee 
such that for every $P\in \Pi_n$,
\be\label{basicmz}
\left|\sum_{k=1}^MW_k|P(z_k)| -\int_\XX |P(x)|d\mu^*(x)\right| \le \epsilon\int_\XX |P(x)|d\mu^*(x).
\ee
{\rm (b)} Let the assumptions of part (a) be satisfied with $\epsilon=1/2$. There exist real numbers $w_1,\cdots,w_M$,  such that $|w_k|\le 2W_k$, $k=1,\cdots,M$, in particular,
\be\label{quadmz}
\sum_{k=1}^M |w_k| \le c\mu^*(\mathbb{B}(\mathbb{K}_{2n},4\delta(\mathbb{K}_{2n},\C_1))),
\ee
and
\be\label{quadrature}
\sum_{k=1}^Mw_kP(z_k) =\int_\XX P(x)d\mu^*(x), \qquad P\in \Pi_n.
\ee
{\rm (c)} Let $\delta>0$, $\C_1$ be a random sample from the probability law $\mu^*_{\mathbb{K}_{2n}}$ given by
$$
\mu^*_{\mathbb{K}_{2n}}(B)=\frac{\mu^*(B\cap \mathbb{K}_{2n})}{\mu^*(\mathbb{K}_{2n})},
$$
and $\epsilon_n=\min(1/n,1/B_{2n})$. 
If 
$$
|\C_1|\ge c\epsilon_n^{-q}\mu^*(\mathbb{K}_{2n})\log\left(\frac{\mu^*(\mathbb{B}(\mathbb{K}_{2n}, \epsilon_n))}{\delta\epsilon_n^q}\right),
$$
then the statements (a) and (b) hold with $\mu^*_{\mathbb{K}_{2n}}$-probability exceeding $1-\delta$. 
\end{theorem}

In order to prove Theorem~\ref{theo:mztheo}, we first recall the following theorem \cite[Theorem~5.1]{mhaskar2020dimension}, applied to our context. 
The statement of \cite[Theorem~5.1]{mhaskar2020dimension} seems to require that $\mu^*$ is a probability measure, but this fact is not required in the proof. 
It is required only that $\mu^*(\mathbb{B}(x,r))\ge cr^q$ for $0<r\le 1$.

\begin{theorem}\label{theo:partition}
Let 
$\tau$ be a positive measure supported on a compact subset of $\XX$,  $\epsilon>0$, $\mathcal{A}$ be a maximal $\epsilon$-distinguishable subset of $\mathsf{supp}(\tau)$, and $K=\BB(\mathcal{A},2\epsilon)$.
Then there exists a subset $\C\subseteq \mathcal{A}\subseteq \mathsf{supp}(\tau)$ and a partition $\{Y_y\}_{y\in\C}$ of $K$ with each of the following properties.
\begin{enumerate}
\item (\textbf{volume property}) For $y\in\C$, $Y_y\subseteq \BB(y,18\epsilon)$,  $(\kappa_1/\kappa_2)7^{-q}\epsilon^q\le \mu^*(Y_y)\le \kappa_2(18 \epsilon)^q$, and\\ 
$\tau(Y_y)\ge  (\kappa_1/\kappa_2)19^{-q}\min_{y\in\mathcal{A}}\tau(\BB(y,\epsilon))>0$.
\item (\textbf{density property}) $\eta(\C)\ge \epsilon$, $\delta(K,\C)\le 18\epsilon$.
\item (\textbf{intersection property})  Let $K_1\subseteq K$ be a compact subset. Then 
$$
\left|\{y\in \C : Y_y\cap K_1 \not=\emptyset\}\right| \le (\kappa_2^2/\kappa_1)(133)^qH_\epsilon(K_1).
$$
\end{enumerate}
\end{theorem}

\noindent\textsc{Proof of Theorem~\ref{theo:mztheo}} (a), (b).\\

We observe first that it is enough to prove this theorem for sufficiently large values of $n$. 
In view of Proposition~\ref{prop:rangeineq}, we may choose $n$ large enough so that for any $P\in\Pi_n$,
\be\label{pf6eqn1}
\|P\|_{1,\mu^*,\XX\setminus \mathbb{K}_{2n}}\le n^{-S}\|P\|_1\le (\epsilon/3)\|P\|_1.
\ee
In this proof, we will write $\delta=\delta(\mathbb{K}_{2n},\C_1)$, so that $\mathbb{K}_{2n} \subset \mathbb{B}(\C_1,\delta)$.  
We use Theorem~\ref{theo:partition} with  $\tau$ to be the measure associating the mass $1$ with each element of $\C_1$, and $\delta$ in place of $\epsilon$. 
If $\mathcal{A}$ is a maximal $\delta$-distinguished subset of $\C_1$, then we denote in this proof, $K= \mathbb{B}(\mathcal{A}, 2\delta)$ and observe that $\mathbb{K}_{2n} \subset \mathbb{B}(\C_1,\delta) \subset K\subset \mathbb{B}(\mathbb{K}_{2n},4\delta)$.
We obtain a partition $\{Y_y\}$ of $K$ as in Theorem~\ref{theo:partition}.
The volume property implies that each $Y_y$ contains at least one element of $\C_1$. 
We construct a subset $\C$ of $\C_1$ by choosing exactly one element of $Y_y\cap\C_1$ for each $y$. We may then re-index $\C_1$, so that without loss of generality, $\C=\{z_1,\cdots,z_N\}$ for some $N\le M$, and re-index $\{Y_y\}$ as $\{Y_k\}$, so that $z_k\in Y_k$, $k=1,\cdots,N$.  
To summarize, we have a subset $\{z_1,\cdots,z_N\}\subseteq\C_1$, and a partition $\{Y_k\}_{k=1}^N$ of $K\supset \mathbb{K}_{2n}$ such that each $Y_k\subset \mathbb{B}(z_k, 36\delta)$ and $\mu^*(Y_k)\sim \delta^q$.
In particular, for any $P\in\Pi_n$, 
\be\label{pf6eqn2}
\|P\|_1-\|P\|_{1,\mu^*,K}\le (\epsilon/3)\|P\|_1.
\ee
We now let $W_k=\mu^*(Y_k)$, $k=1,\cdots,N$, and $W_k=0$, $k=N+1,\cdots,M$. 

The next step is to prove that if $\delta\le c(\epsilon)\min(1/n, 1/B_{2n})$, then 
\be\label{pf6eqn3}
\sup_{y\in \XX}\sum_{k=1}^N \int_{Y_k}|\Phi_{2n}(z_k,y)-\Phi_{2n}(x,y)|d\mu^*(x)\le 2\epsilon/3.
\ee
In this part of the proof, the constants denoted by $c_1,c_2,\cdots$ will retain their value until \eref{pf6eqn3} is proved.
Let $y\in\XX$. 
We let $r\ge \delta$ to be chosen later, and write in this proof, $\mathcal{N}=\{k : \mathsf{dist}(y, Y_k) <r\}$, $\mathcal{L}=\{k : \mathsf{dist}(y, Y_k) \ge r\}$ and for $j=0,1,\cdots$, $\mathcal{L}_j=\{k :  2^jr\le \mathsf{dist}(y, Y_k) <2^{j+1} r\}$. 
Since $r\ge\delta$, and each $Y_k\subset \mathbb{B}(z_k,36\delta)$, there are at most $c_1(r/\delta)^q$ elements in $\mathcal{N}$. 
Using the Bernstein-Lipschitz condition and the fact that $\|\Phi_{2n}(\circ,y)\|_\infty\le c_2n^q$, we deduce that
\be\label{pf6eqn4}
\sum_{k\in \mathcal{N}} \int_{Y_k}|\Phi_{2n}(z_k,y)-\Phi_{2n}(x,y)|d\mu^*(x)\le c_3\mu^*(Y_k)n^qB_{2n}\delta (r/\delta)^q\le c_3\mu^*(\mathcal{B}(z_k,36\delta))n^qB_{2n}\delta (r/\delta)^q \le c_4(nr)^qB_{2n}\delta.
\ee
Next, since $\mu^*(Y_k)\sim \delta^q$, we see that the number of elements in each $\mathcal{L}_j$ is $\sim (2^jr/\delta)^q$. Using Proposition~\ref{prop:kernloc} and the fact that $S>q$, we deduce that
if $r\ge 1/n$, then
\be\label{pf6eqn5}
\begin{aligned}
\sum_{k\in \mathcal{L}} \int_{Y_k}&|\Phi_{2n}(z_k,y)-\Phi_{2n}(x,y)|d\mu^*(x)=\sum_{j=0}^\infty \sum_{k\in \mathcal{L}_j}\int_{Y_k}|\Phi_{2n}(z_k,y)-\Phi_{2n}(x,y)|d\mu^*(x)\nonumber\\
&\le c_5n^q(nr)^{-S}\sum_{j=0}^\infty2^{-jS}\left\{\sum_{k\in \mathcal{L}_j}\mu^*(Y_k)\right\}\le c_6 (nr)^{q-S}.
\end{aligned}
\ee
Since $S>q$, we may choose $r\sim_\epsilon n$ such that $c_6 (nr)^{q-S}\le \epsilon/3$, and then require $\delta\le \min(r, c_7(\epsilon)/B_{2n})$ so that in \eref{pf6eqn4}, $c_4(nr)^qB_{2n}\delta\le \epsilon/3$.
Then \eref{pf6eqn4} and \eref{pf6eqn5} lead to \eref{pf6eqn3}. The proof of \eref{pf6eqn3} being completed, we resume the constant convention as usual.

Next, we observe that for any $P\in\Pi_n$,
$$
P(x)=\int_\XX P(y)\Phi_{2n}(x,y)d\mu^*(y), \qquad x\in\XX.
$$
Therefore, we conclude using \eref{pf6eqn3} that
\be\label{pf6eqn7}
\begin{aligned}
\left|\sum_{k=1}^N\mu^*(Y_k)|P(z_k)| \right.&-\left.\int_K |P(x)|d\mu^*(x)\right| \nonumber\\
&=\left|\sum_{k=1}^N\int_{Y_k}\left(|P(z_k)| - |P(x)|\right)d\mu^*(x)\right|\le \sum_{k=1}^N\int_{Y_k}|P(z_k)-P(x)|d\mu^*(x)\nonumber\\
&\le \sum_{k=1}^N\int_{Y_k}\left|\int_\XX P(y)\left\{\Phi_{2n}(z_k,y)-\Phi_{2n}(x,y)\right\}d\mu^*(y)\right|d\mu^*(x)\nonumber\\
&\le \int_\XX |P(y)|\left\{\sum_{k=1}^N \int_{Y_k}|\Phi_{2n}(z_k,y)-\Phi_{2n}(x,y)|d\mu^*(x)\right\}d\mu^*(y)\le (2\epsilon/3)\int_\XX |P(y)|d\mu^*(y).
\end{aligned}
\ee
Together with \eref{pf6eqn2}, this leads to \eref{basicmz}.
From the definition of $W_k=\mu^*(Y_k)$, $k=1,\cdots, N$, $W_k\le c\delta^q$, and $\sum_{k=1}^N W_k =\mu^*(K)=\mu^*(\mathbb{B}(\mathbb{K}_{2n},4\delta))$. Since $W_k=0$ if $k\ge N+1$, we have now proved \eref{mzwtest}, and thus completed the proof of part (a).

Having proved part (a), the proof of part (b) is by now a routine application of the Hahn-Banach theorem (cf. \cite{indiapap, mnw1, frankbern, modlpmz}).
We apply part (a) with $\epsilon=1/2$. 
Continuing the notation in the proof of part (a), we then have
\be\label{pf6eqn8}
(1/2)\|P\|_1\le \sum_{k=1}^N W_k|P(z_k)|\le (3/2)\|P\|_1, \qquad P\in\Pi_n.
\ee
We now equip $\RR^N$ with the norm $\tn(a_1,\cdots,a_N)\tn=\sum_{k=1}^N W_k|a_k|$, consider the sampling operator $\mathcal{S}:\Pi_n\to \RR^N$ given by $\mathcal{S}(P)=(P(z_1),\cdots, P(z_N))$, let $V$ be the range of this operator, and define a linear functional $x^*$ on $V$ by $x^*(\mathcal{S}(P))=\int_\XX Pd\mu^*$. 
The estimate \eref{pf6eqn8} shows that the norm of this functional is $\le 2$. 
The Hahn-Banach theorem yields a norm-preserving extension $X^*$ of $x^*$ to  $\RR^N$, which in turn, can be identified with a vector $(w_1,\cdots, w_N)\in\RR^N$. 
We set $w_k=0$ if $k\ge N+1$.
Formula \eref{quadrature} then expresses the fact that $X^*$ is an extension of $x^*$. 
The preservation of norms shows that $|w_k|\le 2W_k$ if $k=1,\cdots, N$, and it is clear that for $k=N+1,\cdots, M$, $|w_k|=0=W_k$. 
This completes the proof of part (b). \qed

Part (c) of Theorem~\ref{theo:mztheo} follows immediately from the first two parts and the following lemma.

\begin{lemma}\label{lemma:probcover}
Let $\nu^*$ be a probability measure on $\XX$, $K\subset\mathsf{supp}(\nu^*)$ be a compact set.
Let $\epsilon, \delta\in (0,1]$, $\C$ be a maximal $\epsilon/2$-distinguished subset of $K$, and $\nu_\epsilon=\min_{x\in\C}\nu^*(\mathbb{B}(x,\epsilon/2))$.
If 
$$M\ge c\nu_\epsilon^{-1}\log\left(c_1\mu^*(\mathbb{B}(K,\epsilon))/(\delta\epsilon^q)\right),
$$
and $\{z_1,\cdots,z_M\}$ be random samples from the probability law $\nu^*$ then
\be\label{coverprob}
\mathsf{Prob}_{\nu^*}\left(\{\delta(K,\{z_1,\cdots,z_M\})> \epsilon\}\right)\le \delta.
\ee
\end{lemma}

\begin{Proof}\ % of Lemma~\ref{lemma:probcover}
If $\delta(K,\{z_1,\cdots,z_M\})>\epsilon$, then there exists at least one $x\in \C$ such that $\mathbb{B}(x,\epsilon/2)\cap \{z_1,\cdots,z_M\}=\emptyset$.
For every $x\in \C$, $p_x=\nu^*(\mathbb{B}(x,\epsilon/2))\ge \nu_\epsilon$.
We consider the random variable $z_j$ to be equal to $1$ if $z_j\in \mathbb{B}(x,\epsilon/2)$, and $0$ otherwise.
Using \eref{chernoffbd} with $t=1$, we see that 
$$
\mathsf{Prob}\left(\mathbb{B}(x,\epsilon/2)\cap \{z_1,\cdots,z_M\}=\emptyset\right) \le \exp(-Mp_x/2)\le \exp(-cM\nu_\epsilon).
$$
Since $|\C| \le c_1\mu^*(\mathbb{B}(K,\epsilon))/\epsilon^q$, 
$$
\mathsf{Prob}\left(\{\delta(K,\{z_1,\cdots,z_M\})>\epsilon\}\right) \le c_1\frac{\mu^*(\mathbb{B}(K,\epsilon))}{\epsilon^q}\exp(-cM\nu_\epsilon).
$$
We set the right hand side above to $\delta$ and solve for $M$ to prove the lemma.
\end{Proof}

\bhag{Proofs of the results in Section~\ref{bhag:mainresults}.}\label{bhag:pfsect}

We assume the set-up as in Section~\ref{bhag:mainresults}.
Our first goal is to prove the following theorem.

\begin{theorem}\label{theo:probapprox}
Let  $\tau$, $\nu^*$,  $\mathcal{F}$, $f$ be as described Section~\ref{bhag:mainresults}. We assume the Bernstein-Lipschitz condition. 
Let $0<\delta<1$. 
We assume further that $|\mathcal{F}(y,\epsilon)|\le 1$ for all $y\in\XX$, $\epsilon\in\Omega$. There exist constants $c_1, c_2$, such that if $M\ge c_1n^q\tn\nu^*\tn_{R,0}\log(cnB_n/\delta)$, and $\{(y_1,\epsilon_1), \cdots, (y_M, \epsilon_M)\}$ is a random sample from $\tau$, then
\be\label{probapprox}
\mathsf{Prob}_{\nu^*}\left(\left\{\left\|\frac{1}{M}\sum_{j=1}^M \mathcal{F}(y_j,\epsilon_j)\Phi_n(\circ,y_j)-\sigma_n(\nu^*;f)\right\|_\infty \ge c_3\sqrt{\frac{n^q\tn\nu^*\tn_{R,0}\log(cnB_n\tn\nu^*\tn_{R,0}/\delta)}{M}}\right\}\right)\le \frac{\delta}{\tn\nu^*\tn_{R,0}}.
\ee
\end{theorem}

In order to prove this theorem, we record an observation. The following lemma is an immediate corollary of the Bernstein-Lipschitz condition and Proposition~\ref{prop:rangeineq}.
\begin{lemma}\label{lemma:normingset}
Let the Bernstein-Lipschitz condition be satisfied. Then for every $n>0$ and $\epsilon>0$, there exists a finite set $\C_{n,\epsilon}\subset \mathbb{K}_{2n}$ such that $|\C_{n,\epsilon}|\le cB_n^q\epsilon^{-q}\mu^*(\mathbb{B}(\mathbb{K}_{2n},\epsilon))$ and
for any $P\in\Pi_n$,
\be\label{normingset}
\left|\max_{x\in \C_{n,\epsilon}}|P(x)|-\|P\|_\infty\right| \le \epsilon\|P\|_\infty.
\ee
\end{lemma}

\noindent\textsc{Proof of Theorem~\ref{theo:probapprox}.}\\

Let $x\in\XX$. 
We consider the random variables
$$
Z_j=\mathcal{F}(y_j,\epsilon_j)\Phi_n(x,y_j), \qquad j=1,\cdots,M.
$$
Then in view of \eref{iterexp}, $\mathbb{E}_\tau(Z_j)=\sigma_n(\nu^*;f)(x)$ for every $j$. Further, Proposition~\ref{prop:kernloc} shows that for each $j$, $|Z_j|\le cn^q$. Using \eref{philpnorm} with $\nu^*$ in place of $\nu$, $N=n$, $d=0$, we see that for each $j$,
$$
\int_{\XX\times\Omega}|Z_j|^2d\tau \le \int_\XX |\Phi_n(x,y)|^2d\nu^*(y) \le cn^q\tn\nu^*\tn_{R,0}.
$$
Therefore, Bernstein concentration inequality \eref{bernstein_concentration} implies that for any $t\in (0,1)$,
\be\label{pf5eqn1}
\mathsf{Prob}\left(\left\{\left|\frac{1}{M}\sum_{j=1}^M \mathcal{F}(y_j,\epsilon_j)\Phi_n(x,y_j)-\sigma_n(\nu^*;f)(x)\right|\ge t/2\right\}\right)\le 2\exp\left(-c\frac{t^2M}{n^q\tn\nu^*\tn_{R,0}}\right);
\ee

We now note that $Z_j$, $\sigma_n(\nu^*;f)$ are all in $\Pi_n$. Taking a finite set $\C_{n,1/2}$ as in Lemma~\ref{lemma:normingset}, so that $|\C_{n,1/2}|\le cB_n^q\mu^*(\mathbb{B}(\mathbb{K}_{2n},1/2))\le c_1n^cB_n^q$, we deduce that
$$
\max_{x\in \C_{n,1/2}}\left|\frac{1}{M}\sum_{j=1}^M \mathcal{F}(y_j,\epsilon_j)\Phi_n(x,y_j)-\sigma_n(\nu^*;f)(x)\right|\ge (1/2)\left\|\frac{1}{M}\sum_{j=1}^M \mathcal{F}(y_j,\epsilon_j)\Phi_n(\circ,y_j)-\sigma_n(\nu^*;f)\right\|_\infty.
$$
Then \eref{pf5eqn1} leads to
\be\label{pf5eqn2}
\mathsf{Prob}\left(\left\{\left\|\frac{1}{M}\sum_{j=1}^M \mathcal{F}(y_j,\epsilon_j)\Phi_n(x,y_j)-\sigma_n(\nu^*;f)(x)\right\|_\infty\ge t\right\}\right)\le c_1B_n^qn^c\exp\left(-c_2\frac{t^2M}{n^q\tn\nu^*\tn_{R,0}}\right).
\ee
We set the right hand side above equal to $\delta/\tn\nu^*\tn_{R,0}$ and solve for $t$ to obtain \eref{probapprox} (with different values of $c, c_1, c_2$).
\qed

Before starting to prove results regarding eignets, we first record the continuity and smoothness of a ``smooth kernel'' $G$ as defined in Definition~\ref{def:eignet}.

\begin{prop}\label{prop:eignetbasic}
If $G$ is a smooth kernel, then $(x,y)\mapsto W(y)G(x,y)$ is in $C_0(\XX\times\XX)\cap L^1(\mu^*\times\mu^*;\XX\times\XX)$. 
Further, for any $p$, $1\le p \le \infty$, and $\Lambda\ge 1$,
\be\label{eignetapprox}
 \sup_{x\in\XX}\left\|W(\circ)G(x,\circ)-\sum_{k:\lambda_k<\Lambda}b(\lambda_k)\phi_k(x)\phi_k(\circ)\right\|_p\le c_1\Lambda^cb(\Lambda).
\ee
In particular, for every $x,y\in\XX$, $W(\circ)G(x,\circ)$ and $W(y)G(\circ,y)$ are in $C^\infty$.
\end{prop}

\begin{Proof}\ % of Proposition~\ref{prop:eignetbasic}.
Let $b$ be the smooth mask corresponding to $G$. For any $S\ge 1$, $b(n)\le cn^{-S}b(n/B^*)\le cn^{-S}b(0)$. 
Thus, $b$ itself is fast decreasing. 
Next, let $r>0$. Then  remembering that $B^*\ge 1$ and $b$ is non-increasing, we obtain that for $S>0$, $b(B^*\Lambda u)\le c(\Lambda u)^{-S-r-1}b(\Lambda u)$, and
\be\label{pf7eqn1}
 \int_\Lambda^\infty t^{r}b(t)dt =(B^*\Lambda)^{r+1}\int_{1/B^*}^\infty u^rb(B^*\Lambda u)du \le c\Lambda^{-S}\int_{1/B^*}^\infty u^{-S-1}b(\Lambda u)du\le c\Lambda^{-S}\int_1^\infty u^{-S-1}b(\Lambda u)du\le c\Lambda^{-S}b(\Lambda).
\ee
In this proof, let $s(t)=\sum_{k: \lambda_k<t}\phi_k(x)^2$, so that $s(t)\le ct^q$, $t\ge 1$. 
If $\Lambda\ge 1$, then integrating by parts, we deduce (remembering that $b$ is non-increasing) that for any $x\in\XX$,
\be\label{pf7eqn2}
\begin{aligned}
\sum_{k: \lambda_k\ge \Lambda}b(\lambda_k)\phi_k(x)^2&=\int_\Lambda^\infty b(t)ds(t) = -b(\Lambda)s(\Lambda)-\int_\Lambda^\infty s(t)db(t)\\
&\le c_1\left\{\Lambda^qb(\Lambda)-\int_\Lambda^\infty t^qdb(t)\right\}\le c_2\left\{\Lambda^qb(\Lambda)+\int_\Lambda^\infty t^{q-1}b(t)dt\right\}\\
&\le c_3\Lambda^qb(\Lambda).
\end{aligned}
\ee
Using Schwarz inequality, we conclude that
\be\label{pf7eqn3}
\sup_{x,y\in\XX}\sum_{k: \lambda_k\ge \Lambda}b(\lambda_k)|\phi_k(x)\phi_k(y)| \le c_3\Lambda^qb(\Lambda).
\ee
 In particular, since $b$ is fast decreasing, $W(\circ)G(x,\circ)\in C_0(\XX)$ (and in fact, $W(y)G(x,y)\in C_0(\XX\times\XX)$) and \eref{eignetapprox} holds with $p=\infty$.
Next, for any $j\ge 0$, essential compactness implies that
$$
\int_{\XX\setminus\mathbb{K}_{2^{j+1}\Lambda }}\left(\sum_{k: \lambda_k\in [2^j\Lambda, 2^{j+1}\Lambda)} b(\lambda_k)\phi_k(y)^2\right)^{1/2}d\mu^*(y)
\le c\Lambda^{-S-q}b(2^j\Lambda)^{1/2}.
$$
So, there exists $r\ge q$ such that
$$
\begin{aligned}
\int_\XX \left(\sum_{k: \lambda_k\in [2^j\Lambda, 2^{j+1}\Lambda)}\right.& \left. b(\lambda_k)\phi_k(y)^2\right)^{1/2}d\mu^*(y)\\ &\le \int_{\mathbb{K}_{2^{j+1}\Lambda }}\left(\sum_{k: \lambda_k\in [2^j\Lambda, 2^{j+1}\Lambda)} b(\lambda_k)\phi_k(y)^2\right)^{1/2}d\mu^*(y)+c\Lambda^{-S-q}b(2^j\Lambda)^{1/2}\\
&\le c\left((2^j\Lambda)^q b(2^j\Lambda)\right)^{1/2}\mu^*(\mathbb{K}_{2^{j+1}\Lambda }) \le c\left((2^j\Lambda)^r b(2^j\Lambda)\right)^{1/2}.
\end{aligned}
$$
Hence, for any $x\in\XX$,
\be\label{pf7eqn4}
\begin{aligned}
\int_\XX \sum_{k: \lambda_k\ge \Lambda}&b(\lambda_k)|\phi_k(x)\phi_k(y)| d\mu^*(y)\\
&=\sum_{j=0}^\infty \int_\XX\sum_{k: \lambda_k\in [2^j\Lambda, 2^{j+1}\Lambda)}b(\lambda_k)|\phi_k(x)\phi_k(y)| d\mu^*(y)\\
&\le \sum_{j=0}^\infty\left\{\sum_{k: \lambda_k\in [2^j\Lambda, 2^{j+1}\Lambda)}b(\lambda_k)\phi_k(x)^2\right\}^{1/2}\int_\XX \left(\sum_{k: \lambda_k\in [2^j\Lambda, 2^{j+1}\Lambda)} b(\lambda_k)\phi_k(y)^2\right)^{1/2}d\mu^*(y)\\
&\le c\sum_{j=0}^\infty (2^j\Lambda)^rb(2^j\Lambda) \le c\sum_{j=0}^\infty \int_{2^{j-1}\Lambda}^{2^j\Lambda} t^{r-1}b(t)dt\\
&=c\int_{\Lambda/2}^\infty t^{r-1}b(t)dt \le c\Lambda^{-S}b(\Lambda).
\end{aligned}
\ee
This shows that
\be\label{pf7eqn5}
\sup_{x\in\XX}\left\|\sum_{k: \lambda_k\ge \Lambda}b(\lambda_k)|\phi_k(x)\phi_k(\circ)|\right\|_1\le c\Lambda^{-S}b(\Lambda).
\ee
In view of the convexity inequality:
$$
\|f\|_p\le \|f\|_\infty^{1-1/p}\|f\|_1^{1/p}, \qquad 1<p<\infty,
$$
\eref{pf7eqn3} and \eref{pf7eqn5} lead to
$$
\sup_{x\in\XX}\left\|\sum_{k: \lambda_k\ge \Lambda}b(\lambda_k)|\phi_k(x)\phi_k(\circ)|\right\|_p\le c_1\Lambda^{c}b(\Lambda), \qquad 1\le p\le \infty.
$$
In turn, this implies that $WG(x,\circ)\in L^p$ for all $x\in\XX$, and \eref{eignetapprox} holds.
\end{Proof}

A fundamental fact that relates the kernels $\Phi_n$ and the pre-fabricated eignets $\mathbb{G}_n$'s is the following theorem.

\begin{theorem}\label{theo:eignetkern}
Let $G$ be a smooth kernel, and $\{\nu_n\}$ be an admissible product quadrature measure sequence. 
Then for $1\le p\le\infty$,
$$
\left\{\sup_{x\in\XX}\|\mathbb{G}_n(\nu_{B^*n};x,\circ)-\Phi_n(x,\circ)\|_p\right\}
$$
is fast decreasing.
In particular, for every $S>0$
\be\label{netlocalization}
|\mathbb{G}_n(\nu_{B^*n};x,y)|\le c(S)\left\{\frac{n^{q}}{\max(1, (N\rho(x,y))^S)}+n^{-2S}\right\}.
\ee
\end{theorem}

\begin{Proof}\ % of Theorem~\ref{theo:eignetkern}
Let $x\in\XX$. In this proof, we define $P_n=P_{n,x}$ by $P_n(z)=\sum_{k:\lambda_k<B^*n}b(\lambda_k)\phi_k(x)\phi_k(z)$, $z\in\XX$, and note that $P_n\in\Pi_{B^*n}$. 
In view of Proposition~\ref{prop:eignetbasic}, the expansion in \eref{mercer} converges in $C_0(\XX\times \XX)\cap L^1(\mu^*\times\mu^*;\XX\times\XX)$, so that term-by-term integration can be made to deduce that for $y\in\XX$,
$$
\int_\XX G(x,z)W(z)\mathcal{D}_{G,n}(z,y)d\mu^*(z)=\int_\XX P_n(z)\mathcal{D}_{G,n}(z,y)d\mu^*(z)+ \sum_{k:\lambda_k\ge B^*n}b(\lambda_k)\phi_k(x)\int_\XX \phi_k(z)\mathcal{D}_{G,n}(z,y)d\mu^*(z).
$$ 
By definition, $\mathcal{D}_{G,n}(\circ,y)\in\Pi_n^q$, and hence, each of the summands in the last expression above is equal to $0$. Therefore, recalling that $h(\lambda_k/n)=0$ if $\lambda_k>n$, we obtain
\bea\label{pf8eqn1}
\lefteqn{\int_\XX G(x,z)W(z)\mathcal{D}_{G,n}(z,y)d\mu^*(z)=\int_\XX P_n(z)\mathcal{D}_{G,n}(z,y)d\mu^*(z)}\nonumber\\
&=& \sum_{k:\lambda_k<B^*n} b(\lambda_k)\phi_k(x)\int_\XX \phi_k(z)\mathcal{D}_{G,n}(z,y)d\mu^*(z)\nonumber\\
&=& \sum_{k:\lambda_k<B^*n} b(\lambda_k)\phi_k(x)h(\lambda_k/n)b(\lambda_k)^{-1}\phi_k(y)=\sum_k h(\lambda_k/n)\phi_k(x)\phi_k(y)\nonumber\\
&=&\Phi_n(x,y).
\eea
Since $\mathcal{D}_{G,n}(z,\circ)\in\Pi_n\subset \Pi_{B^*n}$, and $\nu_{B^*n}$ is an admissible product quadrature measure of order $B^*n$, this implies that
\be\label{pf8eqn2}
\Phi_n(x,y)=\int_\XX P_n(z)\mathcal{D}_{G,n}(z,y)d\nu_{B^*n}(z), \qquad y\in\XX.
\ee
Therefore, for $y\in\XX$,
$$
\mathbb{G}_n(\nu_{B^*n};x,y)-\Phi_n(x,y)=\int_\XX\left\{W(z)G(x,z)-P_n(z)\right\}\mathcal{D}_{G,n}(z,y)d\nu_{B^*n}(z).
$$
Using Proposition~\ref{prop:eignetbasic} (used with $\Lambda=B^*n$) and the fact that $\{|\nu_{B^*n}|(\XX)\}$ has polynomial growth, we deduce that
\be\label{pf8eqn3}
\|\mathbb{G}_n(\nu_{B^*n};x,\circ)-\Phi_n(x,\circ)\|_p \le  \left\|W(\circ)G(x,\circ)-P_n\right\|_\infty \sup_{z\in\XX}\|\mathcal{D}_{G,n}(z,\circ)\|_p|\nu_{B^*n}|(\XX)\le c_1n^cb(B^*n)\sup_{z\in\XX}\|\mathcal{D}_{G,n}(z,\circ)\|_p.
\ee
In view of Proposition~\ref{prop:nikolskii} and Proposition~\ref{prop:critical}, we see that
for any $z\in\XX$,
$$
\begin{aligned}
\|\mathcal{D}_{G,n}(z,\circ)\|_p^2&\le c_1n^{2c}\|\mathcal{D}_{G,n}(z,\circ)\|_2^2=c_1n^{2c}\sum_{k: \lambda_k<n} \left(h\left(\lambda_k/n\right)b(\lambda_k)^{-1}\phi_k(z)\right)^2\\
&\le c_1n^{2c}b(n)^{-2}\|\Phi_n(z,\circ)\|_2^2\le c_1n^cb(n)^{-2}\|\Phi_n(z,\circ)\|_1^2\le c_1n^{c}b(n)^{-2}.
\end{aligned}
$$
We now conclude from \eref{pf8eqn3} that
$$
\|\mathbb{G}_n(\nu_{B^*n};x,\circ)-\Phi_n(x,\circ)\|_p\le c_1n^c\frac{b(B^*n)}{b(n)}.
$$
Since $\{b(B^*n)/b(n)\}$ is fast decreasing, this completes the proof.
\end{Proof}\\

The theorems in Section~\ref{bhag:mainresults} all follow from the following basic theorem.

\begin{theorem}\label{theo:eignetprob}
We assume the strong product assumption and the Bernstein-Lipschitz condition. 
With the set-up just described, we have
\be\label{eignetsigmaapprox}
\mathsf{Prob}_{\nu^*}\left(\left\{\left\|\mathcal{G}_n(Y; \mathcal{F})-\sigma_n(f_0f)\right\|_\infty \ge c_3\sqrt{\frac{n^q\tn\nu^*\tn_{R,0}\log(cnB_n\tn\nu^*\tn_{R,0}/\delta)}{|Y|}}\right\}\right)\le \frac{\delta}{\tn\nu^*\tn_{R,0}}.
\ee
In particular, for $f \in X^\infty(\XX)$,  
Then
\be\label{eignetprobapprox}
\mathsf{Prob}_{\nu^*}\left(\left\{\left\|\mathcal{G}_n(Y; \mathcal{F})-f_0f\right\|_\infty \ge c_3\left(\sqrt{\frac{n^q\tn\nu^*\tn_{R,0}\log(cnB_n\tn\nu^*\tn_{R,0}/\delta)}{|Y|}}+E_{n/2}(\infty,f_0f)\right)\right\}\right)\le \frac{\delta}{\tn\nu^*\tn_{R,0}}.
\ee
\end{theorem}

\begin{Proof}\ % of Theorem~\ref{theo:eignetprob}
Theorems~\ref{theo:probapprox} and Theorem~\ref{theo:eignetkern} together lead to \eref{eignetsigmaapprox}.
Since $\sigma_n(\nu^*;f)=\sigma_n(f_0f)$, the estimate \ref{eignetprobapprox} follows from Theorem~\ref{theo:goodapprox} used with $p=\infty$.
\end{Proof}\\

\noindent\textsc{Proof of Theorem~\ref{theo:distributed}}.\\
We observe that with the choice of $f_0$ as in this theorem, $\tn\nu^*\tn_{R,0} \le \|f_0\|_\infty \le 1/\mathfrak{m}$. 
Using $\mathfrak{m}\delta$ in place of $\delta$, we obtain 
Theorem~\ref{theo:distributed} directly from Theorem~\ref{theo:eignetprob} by some simple calculations. \qed\\

\noindent\textsc{Proof of Theorem~\ref{theo:densityest}.}\\

This follows directly from Theorem~\ref{theo:eignetprob} 
by choosing $\mathcal{F}\equiv 1$. \qed\\

\noindent\textsc{Proof of Theorem~\ref{theo:loceignet}}.\\

In view of Theorem~\ref{theo:eignetprob}, our assumptions imply that for each $j\ge 0$,
$$
\mathsf{Prob}_{\nu^*}\left(\left\{\left\|\mathcal{G}_{2^j}(Y; \mathcal{F})-\sigma_{2^j}(f_0f)\right\|_\infty\le c2^{-jS}\right\}\right)\le \delta/2^{j+1}.
$$
Consequenty, with probability $\ge 1-\delta$, we have for each $j\ge 1$,
$$
\left\|\mathcal{G}_{2^j}(Y; \mathcal{F})-\mathcal{G}_{2^{j-1}}(Y_j; \mathcal{F})-\tau_j(f_0f)\right\|_\infty\le c2^{-jS}.
$$
Hence, the theorem follows from Theorem~\ref{theo:paleywiener}.
\qed

\appendix
\renewcommand{\theequation}{\Alph{section}.\arabic{equation}}
\bhag{Gaussian upper bound on manifolds}\label{bhag:gaussupbd}
Let $\XX$ be a compact, connected smooth, $q$-dimensional manifold, $g(x)=(g_{i,j}(x))$ be its metric tensor, and $(g^{i,j}(x))$ be the inverse of $g(x)$. The Laplace-Beltrami operator on $\XX$ is defined by
$$
\Delta(f)(x)=\frac{1}{\sqrt{|g(x)|}}\sum_{i=1}^n\sum_{j=1}^n\partial_i\left(\sqrt{|g(x)|}\, g^{i,j}(x)\partial_j f\right), 
$$
where $|g|=\mbox{det}(g)$.
The symbol of $\Delta$ is given by
$$
a(x,\xi)=\frac{1}{\sqrt{|g(x)|}}\sum_{i=1}^n\sum_{j=1}^n \left(\sqrt{|g(x)|}\, g^{i,j}(x)\right)\xi_i\xi_j.
$$
Then $a(x,\xi)\ge c|\xi|^2$. 
Therefore, H\"ormander's theorem \cite[Theorem~4.4]{hormander1968spectral}, \cite[Theorem~16.1]{shubin1987pseudodifferential} shows that for $x\in\XX$,
\be\label{spectralfn}
\sum_{\lambda_j<\lambda}\phi_k(x)^2 \le c\lambda^q, \qquad \lambda\ge 1.
\ee
In turn, \cite[Proposition~4.1]{frankbern} implies that
$$
\sum_{k=0}^\infty \exp(-\lambda_k^2t)\phi_k(x)^2\le ct^{-q/2}, \qquad t\in (0,1],\ x\in\XX.
$$
 Then \cite[Theorem~1.1]{grigor1997gaussian} shows that 
 \eref{gaussianbd} is satisfied.

\bhag{Probabilistic estimates}\label{bhag:concentration}

We need the following basic facts from probability theory.
Proposition~\ref{prop:concentration}(a) below is a reformulation of  \cite[Section~2.1, 2.7]{boucheron2013concentration}.
A proof of Proposition~\ref{prop:concentration}(b) below is given in \cite[Eqn~(7)]{hagerup1990guided}.

\begin{prop}\label{prop:concentration}
{\rm (a)} (\textbf{Bernstein concentration inequality}) Let $Z_1,\cdots, Z_M$ be independent real valued random variables such that for each $j=1,\cdots,M$, $|Z_j|\le R$, and $\mathbb{E}(Z_j^2)\le V$. Then for any $t>0$,
\be\label{bernstein_concentration}
\mathsf{Prob}\left( \left|\frac{1}{M}\sum_{j=1}^M (Z_j-\mathbb{E}(Z_j))\right| \ge t\right) \le 2\exp\left(-\frac{Mt^2}{2(V+Rt)}\right).
\ee
{\rm (b)} (\textbf{Chernoff  bound})
Let $M\ge 1$, $0\le p\le 1$, and $Z_1,\cdots, Z_M$ be random variables taking values in $\{0,1\}$, with $\mathsf{Prob}(Z_k=1)=p$. Then for $t \in (0,1]$,
\be\label{chernoffbd}
\mathsf{Prob}\left(\sum_{k=1}^M Z_k \le (1-t)Mp\right) \le \exp(-t^2 Mp/2), \quad \mathsf{Prob}\left(\left|\sum_{k=1}^M Z_k-Mp\right| \ge tMp\right) \le 2\exp(-t^2 Mp/2).
\ee
\end{prop}
%R
%\bibliographystyle{abbrv}
%
%\bibliography{/Users/hrushikesh/Documents/hrushikesh/pctexfiles/hrushikesh}
%\end{document}

\end{document}